\documentclass{article}

\usepackage{PRIMEarxiv}
\usepackage{libertine} 

\usepackage[utf8]{inputenc} 
\usepackage[T1]{fontenc}    
\usepackage{hyperref}       
\usepackage{url}            
\usepackage{booktabs}       
\usepackage{amsfonts}       
\usepackage{nicefrac}       
\usepackage{microtype}      
\usepackage{lipsum}
\usepackage{fancyhdr}       
\usepackage{graphicx}       
\usepackage{multirow}
\usepackage{tabularx}
\usepackage{placeins}
\usepackage{longtable}
\graphicspath{{figs/}}     

\title{SmolDocling: An ultra-compact vision-language model for end-to-end multi-modal document conversion}

\author{
  \textbf{Ahmed Nassar}\textsuperscript{1}, \textbf{Andres Marafioti}\textsuperscript{2}, \textbf{Matteo Omenetti}\textsuperscript{1}, \textbf{Maksym Lysak}\textsuperscript{1}, \textbf{Nikolaos Livathinos}\textsuperscript{1}, \\
  \textbf{Christoph Auer}\textsuperscript{1}, \textbf{Lucas Morin}\textsuperscript{1}, \textbf{Rafael Teixeira de Lima}\textsuperscript{1}, \textbf{Yusik Kim}\textsuperscript{1}, \textbf{A. Said Gurbuz}\textsuperscript{1}, \\
  \textbf{Michele Dolfi}\textsuperscript{1}, \textbf{Miquel Farré}\textsuperscript{2}, \textbf{Peter W. J. Staar}\textsuperscript{1} \\
  \\
  IBM Research\textsuperscript{1}, HuggingFace\textsuperscript{2} \\
  \\
  \small{\url{https://huggingface.co/ds4sd/SmolDocling-256M-preview}} \\
  Version 1.0, March 14\textsuperscript{th} 2025
}

\begin{document}
\maketitle

\begin{abstract}
We introduce SmolDocling, an ultra-compact vision-language model targeting end-to-end document conversion. 
Our model comprehensively processes entire pages by generating DocTags, a new universal markup format that captures all page elements in their full context with location. Unlike existing approaches that rely on large foundational models, or ensemble solutions that rely on handcrafted pipelines of multiple specialized models, SmolDocling offers an end-to-end conversion for accurately capturing content, structure and spatial location of document elements in a 256M parameters vision-language model. 
SmolDocling exhibits robust performance in correctly reproducing document features such as code listings, tables, equations, charts, lists, and more across a diverse range of document types including business documents, academic papers, technical reports, patents, and forms — significantly extending beyond the commonly observed focus on scientific papers. 
Additionally, we contribute novel publicly sourced datasets for charts, tables, equations, and code recognition.
Experimental results demonstrate that SmolDocling competes with other Vision Language Models that are up to 27 times larger in size, while reducing computational requirements substantially. The model is currently available, datasets will be publicly available soon.
\end{abstract}
\section{Introduction}
\label{sec:intro}
\FloatBarrier
\begin{figure*}[h!]
    \centering
    \includegraphics*[trim={4.5cm 3.75cm 4.5cm 4.5cm}, clip, width=0.9\textwidth]{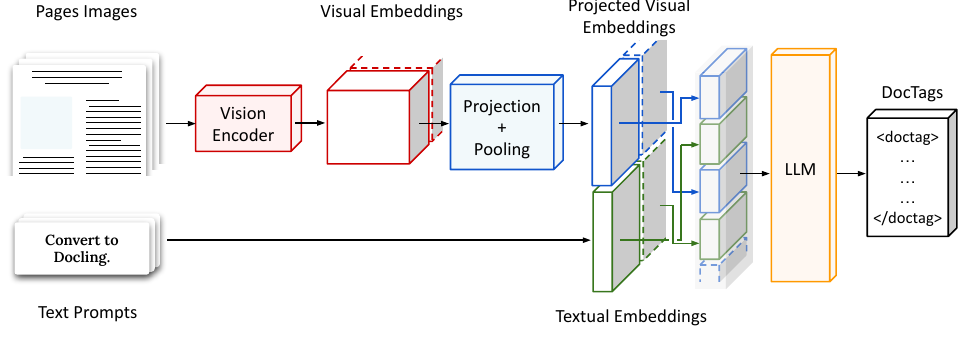}
    \caption{\textbf{SmolDocling/SmolVLM architecture.} SmolDocling converts images of document pages to \textit{DocTags} sequences. First, input images are encoded using a vision encoder and reshaped via projection and pooling. Then, the projected embeddings are concatenated with the text embeddings of the user prompt, possibly with interleaving. Finally, the sequence is used by an LLM to autoregressively predict the \textit{DocTags} sequence.}
    \label{fig:overview}
\end{figure*}
For decades, converting complex digital documents into a structured, machine-processable format has been a significant technical challenge. This challenge primarily stems from the substantial variability in document layouts and styles, as well as the inherently opaque nature of the widely used PDF format, which is optimized for printing rather than semantic parsing. Intricate layout styles and visually challenging elements such as forms, tables, and complex charts can significantly impact the reading order and general understanding of documents.
These problems have driven extensive research and development across multiple domains of computer science. On one hand, sophisticated ensemble systems emerged, which decompose the conversion problem into several sub-tasks (e.g. OCR, layout analysis, table structure recognition, classification) and tackle each sub-task independently. 
Although such systems can achieve high-quality results for many document types while maintaining relatively low computational demands, they are often difficult to tune and generalize.

On the other hand, in the recent past, great interest has developed around large foundational multimodal models that can solve the whole conversion task in one shot, while simultaneously offering flexible querying and parametrization through prompts.
This approach has been made possible by the advent of multimodal pre-training of large vision-language models (LVLMs), which opened up a vast array of opportunities for leveraging diverse data sources, including PDF documents. However, the literature on this topic highlights a significant gap in the availability of high-quality and open-access datasets suitable for training robust multi-modal models for the task of document understanding. 
Furthermore, relying on LVLMs may introduce common issues associated with such models, including hallucinations and the use of significant computational resources, making them impractical from both a quality and cost perspective.

In this work, we outline how we close the gaps left by publicly available datasets and establish a training approach to achieve end-to-end, full-featured document conversion through a vision-language model, which we name \textit{SmolDocling}. Our main contributions can be summarized as follows:

\begin{itemize}
    \item SmolDocling: An ultra-compact VLM for end-to-end document conversion, based on Hugging Face's most recent SmolVLM-256M~\cite{smolvlm}, which is between 5 and 10 times smaller in parameters than comparable VLMs tuned on document understanding tasks. Hence, we reduce the computational complexity to roughly the same order of magnitude as typical representatives of ensemble methods. Uniquely, SmolDocling learns and produces a unified document representation that captures content, structure and spatial location of all elements in a page. 
    \item We augment existing document pre-training datasets with additional feature annotations, and create new datasets for tasks with insufficient coverage in open multi-modal training datasets, including code listings, equations, and chart understanding. Importantly, we construct \textit{full-page} ground truth data that combines key annotation features such as layout, table structure, code listings, charts, and formulas in their natural context, and derive instruction datasets from all training datasets used.
    \item We introduce \textit{DocTags}, a document markup format optimized to efficiently represent the full content and layout characteristics of a document, inspired by ~\cite{tableformer_otsl}.
    \item We evaluate the achieved prediction quality of SmolDocling on several tasks, including comparisons to popular community models.
    
\end{itemize}

\section{Related Work}
\label{sec:related}

\subsection{Large Vision-Language Models (LVLMs)}
With the advent of large language models (LLMs) \cite{raffel2020exploring, openai2023gpt, touvron2023llama, jiang2024mixtral, wei2022chain, anil2023palm, bai2023qwen, granite2024granite, smollm2} characterized by their powerful contextual understanding and adaptability, several approaches have emerged to extend them into large vision-language models. Some proprietary, closed-source LLMs have been adapted into multimodal large language models (MLLMs), such as GPT-4o \cite{openai_hello_gpt4o_2024}, Gemini \cite{team2023gemini}, and Claude 3.5 \cite{anthropic_claude3.5_2024}, demonstrating exceptional capabilities across various modalities, including vision. Open-source methods are actively advancing to match the capabilities of proprietary MLLMs. BLIP-2 \cite{li2023blip} is one of the earliest works combining a vision encoder with a frozen LLM (OPT \cite{zhang2023opt} or FlanT5 \cite{chung2024scaling}) using a lightweight transformer (Q-former \cite{zhang2023vision}). Building upon BLIP-2, MiniGPT-4 \cite{zhu2023minigpt} integrated a frozen visual encoder with a frozen LLM (Vicuna \cite{vicuna2023}) using a Q-Former network and a single projection layer. LLaVA \cite{liu2023llava, liu2023improvedllava} employs a similar architecture but uses a minimal adapter layer. LLaVA-OneVision \cite{li2024llavaB} and LLaVA-NeXT-Interleave \cite{li2024llavainter} build upon LLaVA to support multi-image, higher resolution, and video understanding. For handling high resolutions and text-image compositions and comprehension, InternLM-XComposer introduces a family of models \cite{internlmxcomposer, internlmxcomposer2_5, internlmxcomposer2_4khd, internlmxcomposer2} designed specifically for these tasks. Qwen-VL \cite{bai2023qwen} introduces a position-aware adapter to address efficiency issues caused by the vision encoder generating long image feature sequences. Qwen2.5-VL \cite{qwen2_5vl_tech_report} employs windowed attention with 2D rotary positional embeddings to process native-resolution inputs efficiently and integrates an vision-language merger for dynamic feature compression.

In this work, we build upon the SmolVLM architecture approach~\cite{smolvlm}. Precisely, we adopt SmolVLM-256M, a compact variant of base SmolVLM-2.2B model, inspired by the architecture of Idefics3 \cite{laurençon2024building} which employs the shape-optimized SigLIP encoder as its visual backbone. It applies an aggressive pixel shuffle strategy to compress visual features, reduces the number of image hidden states and introduces new special tokens to separate sub-images and improve tokenization efficiency.

\subsection{State of the Art in Document Understanding}
Several powerful document understanding solutions have emerged on the market in the past decade as commercial cloud offerings on hyperscalers~\cite{google_document_ai, microsoft_ai_document_intelligence, aws_textract, auer2022delivering}, frontier models such as GPT-4o~\cite{openai_hello_gpt4o_2024} or Claude~\cite{anthropic_claude3.5_2024}, or as open-source libraries such as Docling~\cite{livathinos2025doclingefficientopensourcetoolkit}, Grobid~\cite{GROBID}, Marker~\cite{marker}, MinerU~\cite{wang2024mineruopensourcesolutionprecise} or Unstructured~\cite{unstructured_io}. Typical tasks that document understanding encompasses are document classification \cite{xu2020layoutlm, huang2022layoutlmv3}, OCR~\cite{lin2020review}, layout analysis \cite{xu2020layoutlmv2, doclaynet}, table recognition \cite{gte, tableformer_otsl, pub1m}, key-value extraction \cite{huang2022layoutlmv3, kim2021donut, funsd}, chart understanding \cite{liu2022deplot, masry2023unichart, liu2022matcha, han2023chartllama, meng2024chartassisstant, xia2023structchart, xia2024chartx, chen2024onechart, masry2022chartqa, kahou2017figureqa, kantharaj2022chart}, equation recognition \cite{schmitt2024mathnet, paruchuri_texify, deng2017image, yan2021convmath, gervais2404mathwriting}, and more. 

Ensemble systems, such as the aforementioned open-source libraries \cite{livathinos2025doclingefficientopensourcetoolkit, GROBID, marker, wang2024mineruopensourcesolutionprecise}, implement a pipeline in source code, which conditionally applies specialized, single-task models on a given input document, composing the prediction results into a meaningful document representation. Typically, each task entails human-crafted handling of pre-processing and post-processing logic (i.e. setting confidence thresholds for layout detections, matching layout elements to text cells, conditionally applying models).

Conversely, multi-task models aim at providing a single model capable of handling various document-understanding related tasks simultaneously, benefiting from shared context and shared representations learned across these different tasks. OCR-reliant methods, such as LayoutLM \cite{xu2020layoutlm, xu2020layoutlmv2, huang2022layoutlmv3} and UDOP \cite{tang2023unifying}, use text extracted from an external OCR engine, along with image and text bounding box locations, as input. In contrast, OCR-free methods such as Donut~\cite{kim2021donut}, Dessurt~\cite{davis2022end}, DocParser~\cite{rausch2021docparser}, Pix2Struct~\cite{lee2023pix2struct} are transformer-based models trained end-to-end to take an image as input and directly output text. Large Vision Language Models (LVLMs) such as LLaVA~\cite{liu2023improvedllava}, LLaVA-OneVision~\cite{li2024llavaB}, UReader \cite{ye2023ureader}, Kosmos-2~\cite{peng2023kosmos}, and Qwen-VL~\cite{bai2023qwen} all leverage various vision encoders, projection adapters, and LLMs.  While they may use different image patching techniques and instruction-tuning datasets to handle a range of vision tasks, they commonly evaluate their document understanding capabilities on datasets like DocVQA \cite{mathew2021docvqa} and mPLUG-DocOwl 1.5 datasets \cite{hu2024mplugA}, primarily focusing on question answering and reasoning.

Our work focuses primarily on document understanding tasks related to conversion and structural recognition, aiming to represent documents with high accuracy and completeness. The closest works to our approach are Nougat~\cite{nougat}, DocOwl 2~\cite{hu2024mplugB}, GOT~\cite{wei2024general}, and Qwen2.5-VL~\cite{qwen2_5vl_tech_report}. DocOwl 2 and GOT are dedicated to precise conversion and recognition of document structures. On one hand, DocOwl 2 employs a dynamic shape-adaptive cropping module to process high-resolution images, feeding them into a ViT vision encoder. A dedicated module compresses the processed visual features before feeding them into a LLaMa-based LLM. On the other hand, GOT focuses on converting diverse elements—such as text, formulas, molecular diagrams, tables, charts, sheet music, and geometric shapes—into structured formats. It achieves a lightweight architecture by utilizing a SAM vision encoder~\cite{kirillov2023segment}, a linear projection layer, and a Qwen-0.5B LLM. Building on these approaches, our work not only extends the range of tasks but also broadens the types of documents handled, moving beyond a strictly scientific focus to support a more diverse array of document formats. In parallel, Qwen2.5-VL introduces the \textit{Omni-Parsing} strategy, that integrates multiple document elements into a unified HTML-based representation. This format encodes layout bounding box information and reading order for various document elements such as paragraphs and charts.

\section{SmolDocling}
\label{sec:method}
\FloatBarrier
\begin{figure*}[h!]
    \centering
    \includegraphics[width=\textwidth]{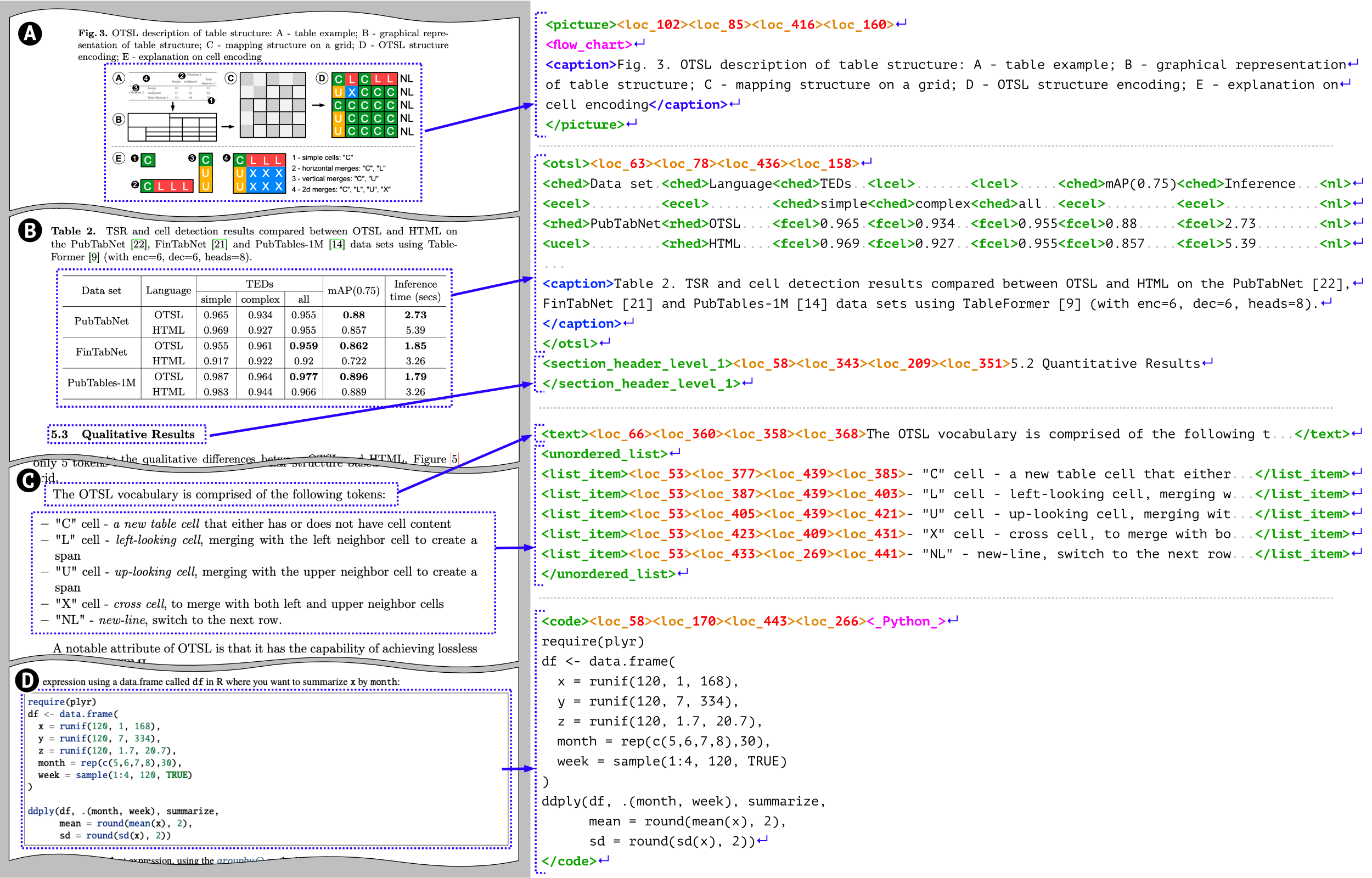}
    \caption{\textbf{DocTags format} describes key features of document elements: type of an element (text, picture, table, code, etc.), location on the page, and content. Nested tags convey additional information: pictures and tables may nest captions, table structure is represented in OTSL tags, lists nest list items, code and pictures carry a classification. All \textit{DocTags} output is taken from SmolDocling predictions, artificial line-breaks and dots (...) were inserted for better readability. A, B, C - snippets of various pages from  \cite{tableformer_otsl}, D - from \cite{pandas}}
    \label{fig:doctags}
\end{figure*}

Building on Hugging Face's SmolVLM~\cite{smolvlm}, SmolDocling converts document page images into sequences representing their content and structure. It produces a unified page representation (DocTags markup) that is optimized for the LLM backbone and interpretable by Docling~\cite{livathinos2025doclingefficientopensourcetoolkit}. This new DocTags format covers the reproduction of captions, charts, forms, code, equations, tables, footnotes, lists, page footers and headers, section headings, and text. SmolDocling both recognizes the type and location of these elements as well as performs OCR within them. Our approach enables full-page conversion with reading-order and hierarchical linking between elements (e.g., linking captions with figures and tables), providing bounding boxes for each component, supporting isolated predictions on individual elements (e.g., cropped tables) and grouping for lists and nested lists. 

\subsection{Model Architecture}

 The model architecture is illustrated in \autoref{fig:overview}. SmolVLM-256M relies on SigLIP base patch-16/512 (93M) as visual backbone and compared to the 2.2B version of the same model, its training data was rebalanced to emphasize document understanding (41\%) and image captioning (14\%), combining The Cauldron~\cite{laurençon2024matters}, Docmatix~\cite{laurençon2024building} datasets with the addition of MathWriting~\cite{gervais2404mathwriting}. It uses lightweight variant of the SmolLM-2 family (135M)~\cite{smollm2} as the language backbone and employs a radical pixel shuffle method that compresses each 512x512 image patch into 64 visual tokens.
 Last but not least, tokenization efficiency was also improved in SmolVLM-256M by increasing the pixel-to-token ratio to 4096 pixels per token and introducing special tokens for sub-image separators.

\subsection{DocTags}
To train VLMs for standardized document conversion, we introduce the \textit{DocTags} markup format, illustrated in \autoref{fig:doctags}. 
Inspired by OTSL~\cite{tableformer_otsl}, \textit{DocTags} define a structured vocabulary of unambiguous tags and rules that explicitly separate textual content from document structure. This disentanglement improves performances of Image-to-Sequence models by minimizing confusions. In contrast, direct conversion to standard formats such as HTML or Markdown is often lossy and ambiguous, lacking explicit structural markup, failing to preserve page layout context, and increasing total tokens count. 
Pieces of text content that represent basic elements in \textit{DocTags} are enclosed in XML-style tags.
Tags define document block types such as: \textit{text, caption, footnote, formula, title, list\_item, page\_footer, page\_header, picture, section\_header, otsl, code, document\_index, etc.} Each element can nest additional location tags encoding its position on the page as a bounding box, represented in DocTags as \texttt{<loc\_\textit{x1}><loc\_\textit{y1}><loc\_\textit{x2}><loc\_\textit{y2}>}.

Special blocks, such as tables and images, nest additional descriptors for captions, tabular structure, or image classes. To encode table structures, we fully include the OTSL vocabulary into \textit{DocTags}. OTSL has the advantage of being minimalist and supporting cell spans and header information. Its tokens are interleaved with text, allowing simultaneous encoding of both table structure and content. In order to promote robust visual-semantic alignment in our document understanding pipeline, we maintain a uniform \textit{DocTags} representation for cropped page elements (e.g., tables, code, equations) that is identical to their full-page counterparts. This consistency ensures that the model can leverage similar visual cues whether an element is presented in isolation or in the context of an entire page, thereby strengthening the learning signal for both computer vision and downstream machine learning tasks.

For a complete list of tags and detailed description of the format, consult the appendix.

\subsection{Model Training}

We employ a curriculum learning approach to progressively align our model for document conversion, ensuring faster convergence. As an initial step, we incorporate DocTags as tokens in our tokenizer. To align the LLM part we freeze the vision encoder and train only the remaining network to adapt it to the new output format which it hasn't seen before. To ensure comprehensive coverage of all DocTags, we maintain a balanced mix of tasks and data types during training. Next, we unfreeze the vision encoder and train the model on our pretraining datasets (see Sec.~\ref{sec:data_pretrain}), along with all task-specific conversion datasets, including tables, code, equations, and charts. Finally, we fine-tune using all available datasets.

\section{Data}
\label{sec:data}

For document understanding tasks, each page image represents the visual layout of textual elements—such as paragraphs, formulas, tables, code, or figures with embedded textual content. Publicly available datasets that unify all these characteristics remain limited, and existing corpora often suffer from small sample sizes or fragmented formats. In this section, we explain how we mitigate these issues in both pre-training and task-specific training scenarios.

\subsection{Pre-training datasets}
\label{sec:data_pretrain}

In the domain of document understanding, currently available open-access document pre-training datasets include OCR-IDL~\cite{ocr-idl}, IIT-CDIP~\cite{iit-cdip}, CCpdf~\cite{CCpdf}, and WordScape~\cite{wordscape}. OCR-IDL and IIT-CDIP, industry-sourced from the 1990s, provide only page images and OCR text. CCpdf and WordScape, derived from Common Crawl, focus on raw PDFs and Word documents, respectively. While WordScape enriches documents with layout, table structure, and topic information via XML markup, its visual variability is limited. Notably, none of these datasets offer rich annotations for all page elements.

\textbf{DocLayNet-PT.} Given the lack of public multimodal document pre-training data with sufficient annotation features, we constructed a new document pre-training dataset named DocLayNet-PT. DocLayNet-PT is a 1.4M pages dataset extracted from the DocFM dataset~\cite{granite2024granite} of unique PDF documents, sourced from CommonCrawl, Wikipedia, and business-related documents. We focused on this dataset to look for documents including visually distinct content such as equations, tables, code and charts with colorful layouts. This dataset was augmented with weak annotations through a series of processing steps. First, we applied PDF parsing to all pages to obtain the text layout including positional information using Docling~\cite{auer2024docling, livathinos2025doclingefficientopensourcetoolkit}. Next, we performed an enrichment step to obtain an enhanced, machine-friendly representation of each page. Specifically, we provide annotations for layout elements, table structure, language, topic, and figure classification on each page in DocTag format. 

\textbf{Docmatix.} To preserve SmolVLM’s original DocVQA capabilities, we adapted the same weak annotation strategy used for DocLayNet-PT and applied it to the 1.3M documents inside Docmatix dataset~\cite{laurençon2024building}. In addition to the 9.5 million DocVQA questions, we introduced another instruction requiring the full conversion of multi-page documents to DocTags.

\subsection{Task-specific Datasets}

\textbf{Layout.} To optimize prediction quality for document layout and table structure, we sampled 76K pages from DocLayNet-PT. Each page underwent human annotation and rigorous quality review, resulting in a high-quality ground-truth dataset. Of these, 60K pages were utilized for fine-tuning, constituting what we refer to as DocLayNet v2.  Additionally, we extracted 63K pages from WordScape, specifically filtering for pages containing both text and tables, and utilized their inherently structure-preserving format as a reliable source of ground truth.
To further enhance the model’s adaptability to diverse layouts, colors, and fonts, we synthetically generated 250K pages using content from Wikipedia, creating the dataset known as SynthDocNet. This synthetic dataset enables the controlled generation of various document types with precise annotations. All datasets share a unified annotation format and class definitions, ensuring consistency and comprehensive document representation.

\textbf{Tables.} For the table structure recognition task we trained our model on public datasets such as PubTables-1M~\cite{pub1m}, FinTabNet~\cite{gte}, WikiTableSet~\cite{icpram23} and tabular information extracted from WordScape documents. We transformed table structure information from the original ground-truth to OTSL format~\cite{tableformer_otsl}. Then, we interleaved each cell tag with ground-truth text. This yields a condensed sequence which represents the structure of the table (including merged cells, and header declaration) along with its textual content.

\textbf{Charts.} For chart reconstruction, several datasets are publicly available~\cite{masry2022chartqa, kahou2017figureqa, kantharaj2022chart, xia2023structchart} that consist of both synthetic and web-scraped charts. However, these datasets are limited by either the quantity of real-world data or the visual diversity of synthetic examples. This has prompted recent studies~\cite{liu2022matcha, han2023chartllama, meng2024chartassisstant, xia2024chartx, xia2023structchart, chen2024onechart, masry2023unichart} to crawl or render additional datasets, which unfortunately are rarely shared publicly. To address these gaps, we generated our own chart dataset comprising four types of charts: line, pie, bar, and stacked bar. Using data from 90,000 tables sourced from the FinTabNet dataset~\cite{gte}, we generated a comprehensive dataset containing 5,000 line charts, 380,000 pie charts, 380,000 bar charts, and 77,000 stacked bar charts. Each data point was rendered using three different visualization libraries (details provided in the Appendix), yielding a total of 2.5 million visually diverse charts.

\textbf{Code.} Technical books and scientific documents frequently include code snippets, making it essential for document-processing models to accurately interpret code structure, particularly indentation, which carries semantic significance in certain programming languages. Several publicly available datasets provide extensive collections of code snippets with permissive licenses in text format~\cite{Kocetkov2022TheStack, puri2021codenet}, typically aggregated from sources like GitHub. However, a common limitation is that these datasets contain only text-based code snippets, lacking any image-based representation. To address this gap, we leveraged LaTeX and Pygments \cite{pygments} to generate visually diverse renderings of code (details provided in the Appendix). Our dataset includes 9.3M million rendered code snippets generated at 120 dpi, covering 56 programming languages.

\textbf{Equations.} Accurately parsing mathematical formulas is crucial for document-processing models, as formulas often contain nuanced structures and semantics critical for scientific understanding. To effectively train our model for formula transcription, we combined publicly available datasets with a custom-rendered dataset. The public datasets \cite{deng2017image, latex-formulas, latex-ocr, gervais2404mathwriting} collectively comprise approximately 730k unique formulas. In addition, we extracted 4.7 million formulas from arXiv by applying a regular expression to the source LaTeX code. 
The extracted formulas underwent a rigorous normalization procedure to ensure the model was trained only on correct and standardized LaTex code. Subsequently, these normalized equations were visually rendered using LaTeX (details provided in the Appendix). Overall, our dataset includes 5.5 million unique formulas, each rendered at a resolution of 120 dpi.

\begin{table}[htb!]
\centering
\resizebox{0.5\linewidth}{!}{
\begin{tabular}{l r l}
\toprule
\textbf{Dataset Name} & \textbf{Size} & \textbf{Type} \\
\midrule
TheCauldron \cite{laurençon2024matters} & 2.63M & \multirow{3}{*}{Multi Task} \\
DocLayNet-PT-Instruct & 1.62M & \\
DocMatix w/ Doc Conversion* \cite{laurençon2024building} & 952K & \\
\midrule
SynthChartNet* & 2.5M & \multirow{5}{*}{Chart} \\
PlotQA \cite{Methani_2020_WACV} & 225K & \\
FigureQA \cite{kahou2017figureqa} & 100K & \\
Unichart \cite{masry2023unichart} & 600K & \\
ChartQA \cite{masry2022chartqa} & 28K & \\
\midrule
DocLayNet-PT & 1.40M & \multirow{6}{*}{Conversion} \\
Doc4MLlaVa - RVLCDIP & 742K & \\
SynthDocNet & 375K & \\
DocLayNet~\cite{doclaynet} & 81K & \\
DocLayNet v2 & 76K & \\
WordScape-PT & 63K & \\
\midrule
PubTable1M \cite{pub1m} & 1M & \multirow{4}{*}{Table} \\
WordScape Tables \cite{wordscape} & 207K & \\
WikiTableSet (EN) \cite{icpram23} & 204K & \\
FinTabNet \cite{gte} & 90K & \\
\midrule
SynthFormulaNet* & 5.5M & Formula \\
\midrule
SynthCodeNet* & 9.3M & Code \\
\bottomrule
\end{tabular}}
\vspace{9pt} 
\caption{\textbf{Datasets.} Overview of the datasets used for the training and evaluation of SmolDocling. Datasets we contribute as part of this publication are denoted with (*).}
\label{tab:datasets}
\end{table}

\subsection{Document Instruction Tuning datasets}
To reinforce the recognition of different page elements and introduce document-related features and no-code pipelines, we leveraged rule-based techniques and the Granite-3.1-2b-instruct \cite{granite2024granite} LLM. Using samples from DocLayNet-PT pages, we generated one instruction by randomly sampling layout elements from a page. These instructions included tasks such as ``Perform OCR at bbox,'' ``Identify page element type at bbox,'' and ``Extract all section headers from the page.'' In addition, we train with Cauldron \cite{laurençon2024matters} to avoid catastrophic forgetting with the amount of conversation datasets introduced.

\begin{table*}[htbp]
\centering
\resizebox{\linewidth}{!}{
\begin{tabular}{lccccccc}
\toprule
Method & Size & \multicolumn{1}{l|}{Edit Distance $\downarrow$} & \multicolumn{1}{l|}{F1-score $\uparrow$} & \multicolumn{1}{l|}{Precision $\uparrow$} & \multicolumn{1}{l|}{Recall $\uparrow$} & \multicolumn{1}{l|}{BLEU $\uparrow$} & \multicolumn{1}{l|}{METEOR $\uparrow$} \\ 
\hline
\multicolumn{8}{l}{\textit{Full-page}} \\
Qwen2.5 VL \cite{qwen2_5vl_tech_report} & 7B & 0.56 & 0.72 & 0.80 & 0.70 & 0.46 & 0.57 \\
GOT \cite{wei2024general} & 580M & 0.61 & 0.69 & 0.71 & 0.73 & 0.48 & 0.59 \\

Nougat (base) \cite{nougat} & 350M & 0.62 & 0.66 & 0.72 & 0.67 & 0.44 & 0.54 \\
\textbf{SmolDocling (Ours)} & 256M & \textbf{0.48} & \textbf{0.80} & \textbf{0.89} & \textbf{0.79} & \textbf{0.58} & \textbf{0.67} \\
\hline
\multicolumn{8}{l}{\textit{Code listings}} \\
\textbf{SmolDocling (Ours)} & 256M & 0.11 & 0.92 & 0.94 & 0.91 & 0.87 & 0.89 \\
\hline
\multicolumn{8}{l}{\textit{Equations}} \\
Qwen2.5 VL \cite{qwen2_5vl_tech_report} & 7B & 0.22 & 0.89 & 0.91 & 0.87 & 0.68 & 0.77 \\
GOT \cite{wei2024general} & 580M & \textbf{0.11} & \textbf{0.95} & \underline{0.95} & \textbf{0.96} & \textbf{0.85} & \textbf{0.91} \\
Nougat (base) \cite{nougat} & 350M & 0.62 & 0.60 & 0.60 & 0.53 & 0.33 & 0.41 \\
\textbf{SmolDocling (Ours)} & 256M & \textbf{0.11} & \textbf{0.95} & \textbf{0.96} & \underline{0.95} & \underline{0.83} & \underline{0.89} \\
\bottomrule
\end{tabular}}
\vspace{6pt}

\caption{\textbf{Structured document text recognition.} We evaluate OCR performance and document formatting on DocLayNet, focusing on textual elements, excluding tables. Text accuracy is measured through multiple metrics (Edit Distance, F1-Score, Precision, Recall, BLEU and METEOR) as proposed in~\cite{nougat}. }
\label{tab:textrecog}
\end{table*}

\section{Experiments}
\label{sec:experiments}

In this section, we assess the performance of SmolDocling on document understanding tasks, and compare the results to previously published models and baselines. Capturing the qualities and defects of SmolDocling and other models in comparable numbers is a non-trivial effort for reasons we outline further below.
Therefore, we report results on standard evaluation tasks and metrics independently (see section~\ref{sec:results_quant}), but also provide a qualitative assessment of output characteristics (see section~\ref{sec:results_qualitative}).

\subsection{Implementation Details} 
Our training was conducted using 64 NVIDIA A100 80GB GPUs, with each epoch requiring 38 hours. The model was trained for a total of 4 epochs using the AdamW optimizer with a learning rate of \(2 \times 10^{-4}\) and \(2 \times 10^{-6}\) for the vision encoder once unfrozen. We also employed gradient clipping to 1.0 for stability and a warmup ratio of 0.03. For inference, using VLLM \cite{kwon2023efficient} on an A100 GPU, SmolDocling achieves a page conversion time of 0.35 seconds while occupying only 0.489 GB of VRAM. The model supports a maximum sequence length of 8,192 tokens and is trained to process up to three pages at a time.

\subsection{Quantitative Results\label{sec:results_quant}}

To enable valid cross-model comparisons with SmolDocling, we addressed multiple dataset and harmonization challenges:

\begin{itemize}
    \item Public evaluation datasets lack comprehensive multi-task annotations with diverse layouts. We augmented DocLayNet \cite{doclaynet} with text content and reading-order to enable evaluation of both layout analysis and text recognition. Table structure, code listing, formula and chart recognition are evaluated on specific datasets.
    \item Different models each follow their own conventions for output markup (e.g., Markdown, HTML, DocTags) and produce annotations which do not fully align in semantics. For example, SmolDocling is trained to produce 14 distinct layout-class tags via \textit{DocTags}, while Qwen2.5-VL combines standard HTML tags and a few explicit class attributes with only partially compatible definitions. GOT and Nougat output Markdown, which can express paragraphs, lists and tables without location annotation. 
    \item Input resolution of page images can affect model performance, but is rarely specified in literature. We standardized at 144 DPI (dots per inch), which reproduces enough detail to be perfectly legible by humans while keeping resource demand for inference reasonable.  
\end{itemize}

\textbf{Text Recognition (OCR).} 
We first evaluate text recognition accuracy of SmolDocling, compared to Nougat, GOT and Qwen2.5-VL (Table \ref{tab:textrecog}), to verify accurate character transcription and reading order. The text-similarity metrics, adopted from~\cite{nougat} require plain formatting without non-textual elements.
For different tasks, we conduct evaluations on diverse content types: full-page documents (from DocLayNet test set) in Markdown format, code snippets (from our SynthCodeNet dataset) in plaintext, and equations (from Im2Latex-230k \cite{eritsyan2023im2latex}) in LaTeX format. For Qwen2.5-VL and SmolDocling, we convert HTML or DocTags output to Markdown while preserving formatting where possible.

Tables are omitted from the full-page output since they do not compare well through Markdown, due to loss of row and column spans. Code snippets must be reproduced with intact line breaks and indentation for accurate scoring.
Formulas are evaluated separately in LaTeX format, which is natively produced by all the models tested in this work. The LaTeX outputs from all models, including the ground truth, undergo the same normalization procedure used for generating the SynthFormulaNet dataset. This ensures consistency and fairness in model comparisons.

SmolDocling significantly outperforms GOT, Nougat and also Qwen2.5-VL, which is 27 times larger, on every metric in full-page transcription (see~\autoref{tab:textrecog}. For the code parsing task, we report results exclusively for SmolDocling, as we introduce this task for the first time, and the other models were not explicitly trained for it.
SmolDocling also demonstrates superior performance compared to Nougat and Qwen2.5-VL in the formula recognition task, closely matching GOT's results. 

\textbf{Layout.}
We evaluate element localization accuracy using the DocLayNet test set (Table \ref{tab:layout}), measuring the ability to reconstruct document representations that support content grounding and visual feature extraction. SmolDocling is compared against Qwen2.5-VL-7b \cite{qwen2_5vl_tech_report}, currently the only other VLM addressing this task. For a fair comparison, we map each model's label outputs and the ground-truth to six compatible classes: Text, Section Heading, List Item, Table, Picture, and Formula.
Overall, SmolDocling outperforms Qwen2.5-VL-7b by a significant margin, however both models score far below a human baseline. This can be partially explained through inherent difficulty of the dataset, as indicated by relatively low human agreement scores. The scores are primarily impacted by moderate recall of element bounding boxes and different degrees of label confusion in both SmolDocling and Qwen2.5-VL. A majority of samples show accurate element bounding boxes as seen in the Appendix.

\begin{table}[htbp]
\centering
\resizebox{0.5\linewidth}{!}{
\begin{tabular}{lccc}
\toprule
                & SmolDocling & Qwen2.5-VL    & Human       \\
\hline
Formula         &  0.267     &  0.059                               & 0.84 \\
Table           &  0.266     &  0.262                               & 0.79   \\
List Item       &  0.296    &  0.090                               & 0.87       \\
Section Header  &  0.289   &  0.129                               & 0.83  \\
Text            &  0.204    &  0.180                               & 0.85   \\ 
Picture         &  0.066    &  0.078                               & 0.70   \\
\hline
Overall         &  0.231    & 0.133                                & 0.82        \\
\bottomrule
\end{tabular}
}
\vspace{9pt}
\caption{\textbf{Layout analysis comparison.} Evaluation of layout analysis task on DocLayNet with a subset of 6 class labels in mAP[0.5:0.95] metric. Additional reference points are provided for an estimate of human performance~\cite{doclaynet}.
}
\label{tab:layout}
\end{table}

\textbf{Table Structure Recognition.}
We evaluate the accuracy of recovering table structures, i.e. shaping columns and rows with the correct text content and spans, through the TEDS metric~\cite{pubtabnet}, once including table text content and once on structure only with ommitted text (see Table~\ref{tab:tablestruct}).
For isolated assessment, we provide cropped table images from FinTabNet \cite{gte} and PubTables-1M \cite{pub1m} test sets, which have poor image quality (72 dpi, compression artifacts). SmolDocling performs competitively against significantly larger models, despite challenges with text transcription from the low-resolution image crops, a limitation likely attributable to insufficient training at such resolutions. Scores for structure only TEDS evaluation are therefore much stronger.

\begin{table}[htb]
\centering
\resizebox{0.6\linewidth}{!}{
\begin{tabular}{lcccc}
\toprule
\textbf{Method} & \textbf{Model Size} & \textbf{FinTabNet} & \textbf{PubTables-1M}\\ \hline
SmolVLM ~\cite{smolvlm} & 2.2B & 0.18  & -  \\
Granite Vision~\cite{granitevision} & 2.98B & 0.54 & -  \\
TableFormer\textdagger~\cite{tableformer_otsl} & 52M & 0.89 & 0.84 \\
\textbf{SmolDocling (Ours)} & 256M & 0.52 (0.81) & 0.65 (0.88) \\

\bottomrule
\end{tabular}}
\vspace{9pt}
\caption{\textbf{Table structure and cell content reconstruction.} Numbers are given in TEDS metric including text content, with structure-only score in brackets where available. We produced results for SmolDocling and TableFormer, other results are reflected here as published. \textdagger TableFormer extracts the content through external OCR.}
\label{tab:tablestruct}
\end{table}

\textbf{Chart Extraction.} In this task, we evaluate the ability to convert cropped sections of charts into datapoint tables. All table ouptut is transformed into HTML and compared against the ground-truth HTML using TEDS score, as reported in \autoref{tab:charts}. We attribute SmolDocling's performance on charts to the variability in ground-truth styles and the differing interpretations of representations across datasets, which we observe leads to inconsistencies in the model's predictions. Despite its significantly smaller size, SmolDocling achieves a competitive TEDs score.

Overall, SmolDocling achieves leading performance across a wide range of document conversion tasks and outperforms significantly larger models in text recognition, layout analysis and structure extraction. Furthermore, the model's high performance in downstream tasks, such as code listing and formula recognition, demonstrates its versatility and real-world applicability. Notably, we establish benchmark results for the code listing recognition, setting a standard for future evaluations.

\subsection{Qualitative Results\label{sec:results_qualitative}}

Several characteristics of SmolDocling output are not evident from task-based metrics alone. In this section, we highlight different output characteristics observed in practice. More illustrated samples can be found in the Appendix.

As presented in Figure~\ref{fig:doctags}, SmolDocling outputs a sequence of DocTags to encode content, structure and layout. A few notable traits beyond straight content reproduction include: 
1) Labeling of page headers and footers, which allows ignoring these repetitive elements in downstream applications. 2) Linking of information, such as captions to tables or pictures, or list elements to parent list through nesting of tags. 3) Preserved line breaks and indentation for code snippets, while line wraps in paragraphs are correctly removed.

Among typical defects found in the output of SmolDocling are missing tags (e.g. absent location tags), malformed structure (such as missing closing tags), and endless repetition of tokens from some point in a page. Both of these phenomena are commonly seen in auto-regressive models and may prevent parsing or rendering the output correctly (see Appendix).
Despite these limitations, SmolDocling demonstrates unique robustness advantages compared to ensemble systems like Docling, since conversion output is inferred in a single pass. In the latter, errors may accumulate throughout the ensemble model pipeline. For example, a mislocated table detection will lead to missing or distorted output of table structure and content, since this depends on correct table localization. In contrast, SmolDocling can reproduce the tables correctly, even with incorrectly determined location.

\begin{table}[]
\centering
\begin{tabular}{lcc}
\hline
Model & Size   & TEDs$\uparrow$ \\
\hline
Phi-3.5-vision & 4B & 0.40 \\
Granite Vision~\cite{granitevision} & 3B & $\mathbf{0.95}$ \\
SmolVLM~\cite{smolvlm} & 2.2B & 0.02 \\
Molmo-E & 1B   & 0.54 \\
\textbf{SmolDocling (Ours)} & 256M & 0.75 \\
\hline
\end{tabular}

\vspace{9pt}
\caption{\textbf{Chart analysis.} We compare SmolDocling to other small-sized VLMs on the chart-to-table task using the TEDS metric. The results in this table are based on the work reported in \cite{granite2024granite}.}
\label{tab:charts}
\end{table}

\section{Conclusion}
\label{sec:conclusion}

In this work, we introduce SmolDocling, an efficient and compact VLM optimized for document conversion while providing rich output representation. We also present a suite of new datasets with a unified format for document conversion, including the novel task of code listing transcription. We identify page element localization as a critical area requiring further refinement, where targeted techniques will significantly enhance performance in future iterations. Our results conclusively demonstrate that smaller models with unified, optimized output formats such as DocTags can effectively compete with significantly larger models, establishing a clear pathway for resource-efficient multi-task document understanding models.

\newpage
{
    \small
    \bibliographystyle{plain}
    \bibliography{11_references.bib}

\begin{thebibliography}{100}

\bibitem{GROBID}
Grobid.
\newblock \url{https://github.com/kermitt2/grobid}, 2008--2025.

\bibitem{smollm2}
Loubna~Ben Allal, Anton Lozhkov, Elie Bakouch, Gabriel~Martín Blázquez, Guilherme Penedo, Lewis Tunstall, Andrés Marafioti, Hynek Kydlíček, Agustín~Piqueres Lajarín, Vaibhav Srivastav, Joshua Lochner, Caleb Fahlgren, Xuan-Son Nguyen, Clémentine Fourrier, Ben Burtenshaw, Hugo Larcher, Haojun Zhao, Cyril Zakka, Mathieu Morlon, Colin Raffel, Leandro von Werra, and Thomas Wolf.
\newblock {SmolLM2: When Smol Goes Big -- Data-Centric Training of a Small Language Model}, 2025.

\bibitem{aws_textract}
{Amazon Web Services}.
\newblock {Amazon Textract}.
\newblock \url{https://aws.amazon.com/textract/}, 2024.
\newblock Accessed: 2024-11-11.

\bibitem{anil2023palm}
Rohan Anil, Andrew~M Dai, Orhan Firat, Melvin Johnson, Dmitry Lepikhin, Alexandre Passos, Siamak Shakeri, Emanuel Taropa, Paige Bailey, Zhifeng Chen, et~al.
\newblock {Palm 2 technical report}.
\newblock {\em arXiv preprint arXiv:2305.10403}, 2023.

\bibitem{anthropic_claude3.5_2024}
Anthropic.
\newblock Claude-3.5.
\newblock \url{https://www.anthropic.com/news/claude-3-5-sonnet}, 2024.
\newblock Accessed: 2024-02-11.

\bibitem{auer2022delivering}
Christoph Auer, Michele Dolfi, Andr{\'e} Carvalho, Cesar~Berrospi Ramis, and Peter~WJ Staar.
\newblock {Delivering Document Conversion as a Cloud Service with High Throughput and Responsiveness}.
\newblock In {\em 2022 IEEE 15th International Conference on Cloud Computing (CLOUD)}, pages 363--373. IEEE, 2022.

\bibitem{auer2024docling}
Christoph Auer, Maksym Lysak, Ahmed Nassar, Michele Dolfi, Nikolaos Livathinos, Panos Vagenas, Cesar~Berrospi Ramis, Matteo Omenetti, Fabian Lindlbauer, Kasper Dinkla, et~al.
\newblock {Docling Technical Report}.
\newblock {\em arXiv preprint arXiv:2408.09869}, 2024.

\bibitem{bai2023qwen}
Jinze Bai, Shuai Bai, Yunfei Chu, Zeyu Cui, Kai Dang, Xiaodong Deng, Yang Fan, Wenbin Ge, Yu~Han, Fei Huang, et~al.
\newblock Qwen technical report.
\newblock {\em arXiv preprint arXiv:2309.16609}, 2023.

\bibitem{qwen2_5vl_tech_report}
Shuai Bai, Keqin Chen, Xuejing Liu, Jialin Wang, Wenbin Ge, Sibo Song, Kai Dang, Peng Wang, Shijie Wang, Jun Tang, Humen Zhong, Yuanzhi Zhu, Mingkun Yang, Zhaohai Li, Jianqiang Wan, Pengfei Wang, Wei Ding, Zheren Fu, Yiheng Xu, Jiabo Ye, Xi~Zhang, Tianbao Xie, Zesen Cheng, Hang Zhang, Zhibo Yang, Haiyang Xu, and Junyang Lin.
\newblock {Qwen2.5-VL Technical Report}, 2025.

\bibitem{ocr-idl}
Ali~Furkan Biten, Rub{\`e}n Tito, Lluis Gomez, Ernest Valveny, and Dimosthenis Karatzas.
\newblock {Ocr-idl: Ocr annotations for industry document library dataset}.
\newblock In {\em European Conference on Computer Vision}, pages 241--252. Springer, 2022.

\bibitem{latex-ocr}
Lukas Blecher.
\newblock {LaTeX-OCR}, 2024.
\newblock Accessed: 2024-11-09.

\bibitem{nougat}
Lukas Blecher, Guillem Cucurull, Thomas Scialom, and Robert Stojnic.
\newblock {Nougat: Neural optical understanding for academic documents}.
\newblock {\em arXiv preprint arXiv:2308.13418}, 2023.

\bibitem{chen2024onechart}
Jinyue Chen, Lingyu Kong, Haoran Wei, Chenglong Liu, Zheng Ge, Liang Zhao, Jianjian Sun, Chunrui Han, and Xiangyu Zhang.
\newblock {OneChart: Purify the Chart Structural Extraction via One Auxiliary Token}.
\newblock {\em arXiv preprint arXiv:2404.09987}, 2024.

\bibitem{chung2024scaling}
Hyung~Won Chung, Le~Hou, Shayne Longpre, Barret Zoph, Yi~Tay, William Fedus, Yunxuan Li, Xuezhi Wang, Mostafa Dehghani, Siddhartha Brahma, et~al.
\newblock Scaling instruction-finetuned language models.
\newblock {\em Journal of Machine Learning Research}, 25(70):1--53, 2024.

\bibitem{pyecharts}
Pyecharts Contributors.
\newblock {pyecharts: Python library for Echarts}, 2024.
\newblock Accessed: 2024-11-06.

\bibitem{pygments}
Pygments Contributors.
\newblock Pygments: Python syntax highlighter, 2024.
\newblock Accessed: 2024-11-06.

\bibitem{davis2022end}
Brian Davis, Bryan Morse, Brian Price, Chris Tensmeyer, Curtis Wigington, and Vlad Morariu.
\newblock End-to-end document recognition and understanding with dessurt.
\newblock In {\em European Conference on Computer Vision}, pages 280--296. Springer, 2022.

\bibitem{deng2017image}
Yuntian Deng, Anssi Kanervisto, Jeffrey Ling, and Alexander~M Rush.
\newblock {Image-to-markup generation with coarse-to-fine attention}.
\newblock In {\em International Conference on Machine Learning}, pages 980--989. PMLR, 2017.

\bibitem{internlmxcomposer2}
Xiaoyi Dong, Pan Zhang, Yuhang Zang, Yuhang Cao, Bin Wang, Linke Ouyang, Xilin Wei, Songyang Zhang, Haodong Duan, Maosong Cao, Wenwei Zhang, Yining Li, Hang Yan, Yang Gao, Xinyue Zhang, Wei Li, Jingwen Li, Kai Chen, Conghui He, Xingcheng Zhang, Yu~Qiao, Dahua Lin, and Jiaqi Wang.
\newblock {InternLM-XComposer2: Mastering Free-form Text-Image Composition and Comprehension in Vision-Language Large Model}.
\newblock {\em arXiv preprint arXiv:2401.16420}, 2024.

\bibitem{internlmxcomposer2_4khd}
Xiaoyi Dong, Pan Zhang, Yuhang Zang, Yuhang Cao, Bin Wang, Linke Ouyang, Songyang Zhang, Haodong Duan, Wenwei Zhang, Yining Li, Hang Yan, Yang Gao, Zhe Chen, Xinyue Zhang, Wei Li, Jingwen Li, Wenhai Wang, Kai Chen, Conghui He, Xingcheng Zhang, Jifeng Dai, Yu~Qiao, Dahua Lin, and Jiaqi Wang.
\newblock {InternLM-XComposer2-4KHD: A Pioneering Large Vision-Language Model Handling Resolutions from 336 Pixels to 4K HD}.
\newblock {\em arXiv preprint arXiv:2404.06512}, 2024.

\bibitem{eritsyan2023im2latex}
Gregory Eritsyan.
\newblock {Im2LaTeX 230K}.
\newblock \url{https://www.kaggle.com/datasets/gregoryeritsyan/im2latex-230k}, 2023.
\newblock Accessed: 2025-03-06.

\bibitem{latex-formulas}
Hugging Face.
\newblock {latex-formulas Dataset}, 2024.
\newblock Accessed: 2024-11-09.

\bibitem{gervais2404mathwriting}
Philippe Gervais, Asya Fadeeva, and Andrii Maksai.
\newblock {Mathwriting: A dataset for handwritten mathematical expression recognition}.
\newblock {\em URL https://arxiv. org/abs/2404.10690}, 2024.

\bibitem{google_document_ai}
{Google Cloud}.
\newblock {Document AI}.
\newblock \url{https://cloud.google.com/document-ai}, 2024.
\newblock Accessed: 2024-11-11.

\bibitem{granite2024granite}
IBM Granite~Team.
\newblock {Granite 3.0 Language Models}, 2024.

\bibitem{han2023chartllama}
Yucheng Han, Chi Zhang, Xin Chen, Xu~Yang, Zhibin Wang, Gang Yu, Bin Fu, and Hanwang Zhang.
\newblock {Chartllama: A multimodal llm for chart understanding and generation}.
\newblock {\em arXiv preprint arXiv:2311.16483}, 2023.

\bibitem{hu2024mplugA}
Anwen Hu, Haiyang Xu, Jiabo Ye, Ming Yan, Liang Zhang, Bo~Zhang, Chen Li, Ji~Zhang, Qin Jin, Fei Huang, et~al.
\newblock mplug-docowl 1.5: Unified structure learning for ocr-free document understanding.
\newblock {\em arXiv preprint arXiv:2403.12895}, 2024.

\bibitem{hu2024mplugB}
Anwen Hu, Haiyang Xu, Liang Zhang, Jiabo Ye, Ming Yan, Ji~Zhang, Qin Jin, Fei Huang, and Jingren Zhou.
\newblock mplug-docowl2: High-resolution compressing for ocr-free multi-page document understanding.
\newblock {\em arXiv preprint arXiv:2409.03420}, 2024.

\bibitem{huang2022layoutlmv3}
Yupan Huang, Tengchao Lv, Lei Cui, Yutong Lu, and Furu Wei.
\newblock {Layoutlmv3: Pre-training for document ai with unified text and image masking}.
\newblock In {\em Proceedings of the 30th ACM International Conference on Multimedia}, pages 4083--4091, 2022.

\bibitem{matplotlib}
J.~D. Hunter.
\newblock {Matplotlib: A 2D graphics environment}.
\newblock {\em Computing in Science \& Engineering}, 9(3):90--95, 2007.

\bibitem{funsd}
Guillaume Jaume, Hazim~Kemal Ekenel, and Jean-Philippe Thiran.
\newblock {Funsd: A dataset for form understanding in noisy scanned documents}.
\newblock In {\em 2019 International Conference on Document Analysis and Recognition Workshops (ICDARW)}, volume~2, pages 1--6. IEEE, 2019.

\bibitem{jiang2024mixtral}
Albert~Q Jiang, Alexandre Sablayrolles, Antoine Roux, Arthur Mensch, Blanche Savary, Chris Bamford, Devendra~Singh Chaplot, Diego de~las Casas, Emma~Bou Hanna, Florian Bressand, et~al.
\newblock Mixtral of experts.
\newblock {\em arXiv preprint arXiv:2401.04088}, 2024.

\bibitem{kahou2017figureqa}
Samira~Ebrahimi Kahou, Vincent Michalski, Adam Atkinson, {\'A}kos K{\'a}d{\'a}r, Adam Trischler, and Yoshua Bengio.
\newblock Figureqa: An annotated figure dataset for visual reasoning.
\newblock {\em arXiv preprint arXiv:1710.07300}, 2017.

\bibitem{kantharaj2022chart}
Shankar Kantharaj, Rixie Tiffany~Ko Leong, Xiang Lin, Ahmed Masry, Megh Thakkar, Enamul Hoque, and Shafiq Joty.
\newblock Chart-to-text: A large-scale benchmark for chart summarization.
\newblock {\em arXiv preprint arXiv:2203.06486}, 2022.

\bibitem{kim2021donut}
Geewook Kim, Teakgyu Hong, Moonbin Yim, Jinyoung Park, Jinyeong Yim, Wonseok Hwang, Sangdoo Yun, Dongyoon Han, and Seunghyun Park.
\newblock Donut: Document understanding transformer without ocr.
\newblock {\em arXiv preprint arXiv:2111.15664}, 7(15):2, 2021.

\bibitem{10.1093/nar/gky1033}
Sunghwan Kim, Jie Chen, Tiejun Cheng, Asta Gindulyte, Jia He, Siqian He, Qingliang Li, Benjamin~A Shoemaker, Paul~A Thiessen, Bo~Yu, Leonid Zaslavsky, Jian Zhang, and Evan~E Bolton.
\newblock {PubChem 2019 update: improved access to chemical data}.
\newblock {\em Nucleic Acids Research}, 47(D1):D1102--D1109, 10 2018.

\bibitem{kirillov2023segment}
Alexander Kirillov, Eric Mintun, Nikhila Ravi, Hanzi Mao, Chloe Rolland, Laura Gustafson, Tete Xiao, Spencer Whitehead, Alexander~C Berg, Wan-Yen Lo, et~al.
\newblock Segment anything.
\newblock In {\em Proceedings of the IEEE/CVF International Conference on Computer Vision}, pages 4015--4026, 2023.

\bibitem{Kocetkov2022TheStack}
Denis Kocetkov, Raymond Li, Loubna Ben~Allal, Jia Li, Chenghao Mou, Carlos Muñoz~Ferrandis, Yacine Jernite, Margaret Mitchell, Sean Hughes, Thomas Wolf, Dzmitry Bahdanau, Leandro von Werra, and Harm de~Vries.
\newblock {The Stack: 3 TB of permissively licensed source code}.
\newblock {\em Preprint}, 2022.

\bibitem{kwon2023efficient}
Woosuk Kwon, Zhuohan Li, Siyuan Zhuang, Ying Sheng, Lianmin Zheng, Cody~Hao Yu, Joseph~E. Gonzalez, Hao Zhang, and Ion Stoica.
\newblock {Efficient Memory Management for Large Language Model Serving with PagedAttention}.
\newblock In {\em Proceedings of the ACM SIGOPS 29th Symposium on Operating Systems Principles}, 2023.

\bibitem{RDKit}
Greg Landrum.
\newblock {RDKit: Open-Source Cheminformatics Software}.
\newblock http://www.rdkit.org/.
\newblock Accessed: 1 January 2023.

\bibitem{laurençon2024building}
Hugo Laurençon, Andrés Marafioti, Victor Sanh, and Léo Tronchon.
\newblock Building and better understanding vision-language models: insights and future directions., 2024.

\bibitem{laurençon2024matters}
Hugo Laurençon, Léo Tronchon, Matthieu Cord, and Victor Sanh.
\newblock What matters when building vision-language models?, 2024.

\bibitem{lee2023pix2struct}
Kenton Lee, Mandar Joshi, Iulia~Raluca Turc, Hexiang Hu, Fangyu Liu, Julian~Martin Eisenschlos, Urvashi Khandelwal, Peter Shaw, Ming-Wei Chang, and Kristina Toutanova.
\newblock Pix2struct: Screenshot parsing as pretraining for visual language understanding.
\newblock In {\em International Conference on Machine Learning}, pages 18893--18912. PMLR, 2023.

\bibitem{iit-cdip}
David Lewis, Gady Agam, Shlomo Argamon, Ophir Frieder, David Grossman, and Jefferson Heard.
\newblock Building a test collection for complex document information processing.
\newblock In {\em Proceedings of the 29th annual international ACM SIGIR conference on Research and development in information retrieval}, pages 665--666, 2006.

\bibitem{li2024llavaB}
Bo~Li, Yuanhan Zhang, Dong Guo, Renrui Zhang, Feng Li, Hao Zhang, Kaichen Zhang, Yanwei Li, Ziwei Liu, and Chunyuan Li.
\newblock {Llava-onevision: Easy visual task transfer}.
\newblock {\em arXiv preprint arXiv:2408.03326}, 2024.

\bibitem{li2024llavainter}
Feng Li, Renrui Zhang, Hao Zhang, Yuanhan Zhang, Bo~Li, Wei Li, Zejun Ma, and Chunyuan Li.
\newblock {Llava-next-interleave: Tackling multi-image, video, and 3d in large multimodal models}.
\newblock {\em arXiv preprint arXiv:2407.07895}, 2024.

\bibitem{li2023blip}
Junnan Li, Dongxu Li, Silvio Savarese, and Steven Hoi.
\newblock Blip-2: Bootstrapping language-image pre-training with frozen image encoders and large language models.
\newblock In {\em International conference on machine learning}, pages 19730--19742. PMLR, 2023.

\bibitem{lin2020review}
Han Lin, Peng Yang, and Fanlong Zhang.
\newblock Review of scene text detection and recognition.
\newblock {\em Archives of computational methods in engineering}, 27(2):433--454, April 2020.

\bibitem{liu2022deplot}
Fangyu Liu, Julian~Martin Eisenschlos, Francesco Piccinno, Syrine Krichene, Chenxi Pang, Kenton Lee, Mandar Joshi, Wenhu Chen, Nigel Collier, and Yasemin Altun.
\newblock Deplot: One-shot visual language reasoning by plot-to-table translation.
\newblock {\em arXiv preprint arXiv:2212.10505}, 2022.

\bibitem{liu2022matcha}
Fangyu Liu, Francesco Piccinno, Syrine Krichene, Chenxi Pang, Kenton Lee, Mandar Joshi, Yasemin Altun, Nigel Collier, and Julian~Martin Eisenschlos.
\newblock Matcha: Enhancing visual language pretraining with math reasoning and chart derendering.
\newblock {\em arXiv preprint arXiv:2212.09662}, 2022.

\bibitem{liu2023improvedllava}
Haotian Liu, Chunyuan Li, Yuheng Li, and Yong~Jae Lee.
\newblock {Improved Baselines with Visual Instruction Tuning}, 2023.

\bibitem{liu2023llava}
Haotian Liu, Chunyuan Li, Qingyang Wu, and Yong~Jae Lee.
\newblock {Visual Instruction Tuning}.
\newblock In {\em NeurIPS}, 2023.

\bibitem{livathinos2025doclingefficientopensourcetoolkit}
Nikolaos Livathinos, Christoph Auer, Maksym Lysak, Ahmed Nassar, Michele Dolfi, Panos Vagenas, Cesar~Berrospi Ramis, Matteo Omenetti, Kasper Dinkla, Yusik Kim, Shubham Gupta, Rafael~Teixeira de~Lima, Valery Weber, Lucas Morin, Ingmar Meijer, Viktor Kuropiatnyk, and Peter W.~J. Staar.
\newblock {Docling: An Efficient Open-Source Toolkit for AI-driven Document Conversion}, 2025.

\bibitem{icpram23}
Nam Ly., Atsuhiro Takasu., Phuc Nguyen., and Hideaki Takeda.
\newblock {Rethinking Image-Based Table Recognition Using Weakly Supervised Methods}.
\newblock pages 872--880, 2023.

\bibitem{tableformer_otsl}
Maksym Lysak, Ahmed Nassar, Nikolaos Livathinos, Christoph Auer, and Peter Staar.
\newblock {Optimized Table Tokenization for Table Structure Recognition}.
\newblock In Gernot~A. Fink, Rajiv Jain, Koichi Kise, and Richard Zanibbi, editors, {\em Document Analysis and Recognition - ICDAR 2023}, pages 37--50, Cham, 2023. Springer Nature Switzerland.

\bibitem{smolvlm}
Andr\'{e}s Marafioti, Orr Zohar, Miquel Farr\'{e}, Merve Noyan, Elie Bakouch, Pedro Cuenca, Cyril Zakka, Loubna Ben~Allal, Anton Lozhkov, Nouamane Tazi, Vaibhav Srivastav, Joshua Lochner, Hugo Larcher, Mathieu Morlon, Lewis Tunstall, Leandro von Werra, and Thomas Wolf.
\newblock {SmolVLM: Redefining small and efficient multimodal models}.
\newblock 2025.

\bibitem{masry2023unichart}
Ahmed Masry, Parsa Kavehzadeh, Xuan~Long Do, Enamul Hoque, and Shafiq Joty.
\newblock Unichart: A universal vision-language pretrained model for chart comprehension and reasoning.
\newblock {\em arXiv preprint arXiv:2305.14761}, 2023.

\bibitem{masry2022chartqa}
Ahmed Masry, Do~Xuan Long, Jia~Qing Tan, Shafiq Joty, and Enamul Hoque.
\newblock Chartqa: A benchmark for question answering about charts with visual and logical reasoning.
\newblock {\em arXiv preprint arXiv:2203.10244}, 2022.

\bibitem{mathew2021docvqa}
Minesh Mathew, Dimosthenis Karatzas, and CV~Jawahar.
\newblock Docvqa: A dataset for vqa on document images.
\newblock In {\em Proceedings of the IEEE/CVF winter conference on applications of computer vision}, pages 2200--2209, 2021.

\bibitem{meng2024chartassisstant}
Fanqing Meng, Wenqi Shao, Quanfeng Lu, Peng Gao, Kaipeng Zhang, Yu~Qiao, and Ping Luo.
\newblock Chartassisstant: A universal chart multimodal language model via chart-to-table pre-training and multitask instruction tuning.
\newblock {\em arXiv preprint arXiv:2401.02384}, 2024.

\bibitem{Methani_2020_WACV}
Nitesh Methani, Pritha Ganguly, Mitesh~M. Khapra, and Pratyush Kumar.
\newblock {PlotQA: Reasoning over Scientific Plots}.
\newblock In {\em The IEEE Winter Conference on Applications of Computer Vision (WACV)}, March 2020.

\bibitem{microsoft_ai_document_intelligence}
{Microsoft Azure}.
\newblock {AI Document Intelligence}.
\newblock \url{https://azure.microsoft.com/en-us/products/ai-services/ai-document-intelligence}, 2024.
\newblock Accessed: 2024-11-11.

\bibitem{Morin_2023_ICCV}
Lucas Morin, Martin Danelljan, Maria~Isabel Agea, Ahmed Nassar, Valery Weber, Ingmar Meijer, Peter Staar, and Fisher Yu.
\newblock {MolGrapher: Graph-based Visual Recognition of Chemical Structures}.
\newblock In {\em Proceedings of the IEEE/CVF International Conference on Computer Vision (ICCV)}, pages 19552--19561, October 2023.

\bibitem{Morin2024}
Lucas Morin, Val{\'e}ry Weber, Gerhard~Ingmar Meijer, Fisher Yu, and Peter W.~J. Staar.
\newblock {PatCID: an open-access dataset of chemical structures in patent documents}.
\newblock {\em Nature Communications}, 15(1):6532, Aug 2024.

\bibitem{openai_hello_gpt4o_2024}
OpenAI.
\newblock {Hello GPT-4O}.
\newblock \url{https://openai.com/index/hello-gpt-4o/}, 2024.
\newblock Accessed: 2024-02-09, 2024-02-11, 2024-02-12.

\bibitem{openai2023gpt}
R~OpenAI.
\newblock Gpt-4 technical report. arxiv 2303.08774.
\newblock {\em View in Article}, 2(5), 2023.

\bibitem{marker}
Vik Paruchuri.
\newblock {Marker: Convert PDF to Markdown Quickly with High Accuracy}.
\newblock https://github.com/VikParuchuri/marker, 2024.

\bibitem{paruchuri_texify}
Vik Paruchuri.
\newblock Texify.
\newblock \url{https://github.com/VikParuchuri/texify}, 2024.
\newblock Accessed: 2024-11-11.

\bibitem{peng2023kosmos}
Zhiliang Peng, Wenhui Wang, Li~Dong, Yaru Hao, Shaohan Huang, Shuming Ma, and Furu Wei.
\newblock Kosmos-2: Grounding multimodal large language models to the world.
\newblock {\em arXiv preprint arXiv:2306.14824}, 2023.

\bibitem{doclaynet}
Birgit Pfitzmann, Christoph Auer, Michele Dolfi, Ahmed~S. Nassar, and Peter Staar.
\newblock {DocLayNet: A Large Human-Annotated Dataset for Document-Layout Segmentation}.
\newblock In {\em Proceedings of the 28th ACM SIGKDD Conference on Knowledge Discovery and Data Mining}, KDD '22, page 3743–3751, New York, NY, USA, 2022. Association for Computing Machinery.

\bibitem{puri2021codenet}
Ruchir Puri, David Kung, Geert Janssen, Wei Zhang, Giacomo Domeniconi, Vladmir Zolotov, Julian Dolby, Jie Chen, Mihir Choudhury, Lindsey Decker, Veronika Thost, Luca Buratti, Saurabh Pujar, Shyam Ramji, Ulrich Finkler, Susan Malaika, and Frederick Reiss.
\newblock {CodeNet: A Large-Scale AI for Code Dataset for Learning a Diversity of Coding Tasks}, 2021.

\bibitem{Pyzer-Knapp2025}
Edward~O. Pyzer-Knapp, Matteo Manica, Peter Staar, Lucas Morin, Patrick Ruch, Teodoro Laino, John~R. Smith, and Alessandro Curioni.
\newblock {Foundation models for materials discovery -- current state and future directions}.
\newblock {\em npj Computational Materials}, 11(1):61, Mar 2025.

\bibitem{Qian2023}
Yujie Qian, Jiang Guo, Zhengkai Tu, Zhening Li, Connor~W. Coley, and Regina Barzilay.
\newblock {MolScribe: Robust Molecular Structure Recognition with Image-to-Graph Generation}.
\newblock {\em Journal of Chemical Information and Modeling}, 63(7):1925--1934, Apr 2023.

\bibitem{raffel2020exploring}
Colin Raffel, Noam Shazeer, Adam Roberts, Katherine Lee, Sharan Narang, Michael Matena, Yanqi Zhou, Wei Li, and Peter~J Liu.
\newblock Exploring the limits of transfer learning with a unified text-to-text transformer.
\newblock {\em Journal of machine learning research}, 21(140):1--67, 2020.

\bibitem{Rajan2023}
Kohulan Rajan, Henning~Otto Brinkhaus, M.~Isabel Agea, Achim Zielesny, and Christoph Steinbeck.
\newblock {DECIMER.ai: an open platform for automated optical chemical structure identification, segmentation and recognition in scientific publications}.
\newblock {\em Nature Communications}, 14(1):5045, Aug 2023.

\bibitem{rausch2021docparser}
Johannes Rausch, Octavio Martinez, Fabian Bissig, Ce~Zhang, and Stefan Feuerriegel.
\newblock Docparser: Hierarchical document structure parsing from renderings.
\newblock In {\em Proceedings of the AAAI Conference on Artificial Intelligence}, volume~35, pages 4328--4338, 2021.

\bibitem{schmitt2024mathnet}
Felix~M Schmitt-Koopmann, Elaine~M Huang, Hans-Peter Hutter, Thilo Stadelmann, and Alireza Darvishy.
\newblock {MathNet: A Data-Centric Approach for Printed Mathematical Expression Recognition}.
\newblock {\em IEEE Access}, 2024.

\bibitem{pub1m}
Brandon Smock, Rohith Pesala, and Robin Abraham.
\newblock {PubTables-1M: Towards comprehensive table extraction from unstructured documents}.
\newblock In {\em 2022 IEEE/CVF Conference on Computer Vision and Pattern Recognition (CVPR)}, pages 4624--4632, 2022.

\bibitem{tang2023unifying}
Zineng Tang, Ziyi Yang, Guoxin Wang, Yuwei Fang, Yang Liu, Chenguang Zhu, Michael Zeng, Cha Zhang, and Mohit Bansal.
\newblock Unifying vision, text, and layout for universal document processing.
\newblock In {\em Proceedings of the IEEE/CVF conference on computer vision and pattern recognition}, pages 19254--19264, 2023.

\bibitem{team2023gemini}
Gemini Team, Rohan Anil, Sebastian Borgeaud, Jean-Baptiste Alayrac, Jiahui Yu, Radu Soricut, Johan Schalkwyk, Andrew~M Dai, Anja Hauth, Katie Millican, et~al.
\newblock Gemini: a family of highly capable multimodal models.
\newblock {\em arXiv preprint arXiv:2312.11805}, 2023.

\bibitem{granitevision}
Granite~Vision Team, Leonid Karlinsky, Assaf Arbelle, Abraham Daniels, Ahmed Nassar, Amit Alfassi, Bo~Wu, Eli Schwartz, Dhiraj Joshi, Jovana Kondic, Nimrod Shabtay, Pengyuan Li, Roei Herzig, Shafiq Abedin, Shaked Perek, Sivan Harary, Udi Barzelay, Adi~Raz Goldfarb, Aude Oliva, Ben Wieles, Bishwaranjan Bhattacharjee, Brandon Huang, Christoph Auer, Dan Gutfreund, David Beymer, David Wood, Hilde Kuehne, Jacob Hansen, Joseph Shtok, Ken Wong, Luis~Angel Bathen, Mayank Mishra, Maksym Lysak, Michele Dolfi, Mikhail Yurochkin, Nikolaos Livathinos, Nimrod Harel, Ophir Azulai, Oshri Naparstek, Rafael~Teixeira de~Lima, Rameswar Panda, Sivan Doveh, Shubham Gupta, Subhro Das, Syed Zawad, Yusik Kim, Zexue He, Alexander Brooks, Gabe Goodhart, Anita Govindjee, Derek Leist, Ibrahim Ibrahim, Aya Soffer, David Cox, Kate Soule, Luis Lastras, Nirmit Desai, Shila Ofek-koifman, Sriram Raghavan, Tanveer Syeda-Mahmood, Peter Staar, Tal Drory, and Rogerio Feris.
\newblock {Granite Vision: a lightweight, open-source multimodal model for enterprise Intelligence}, 2025.

\bibitem{touvron2023llama}
Hugo Touvron, Thibaut Lavril, Gautier Izacard, Xavier Martinet, Marie-Anne Lachaux, Timoth{\'e}e Lacroix, Baptiste Rozi{\`e}re, Naman Goyal, Eric Hambro, Faisal Azhar, et~al.
\newblock {Llama: Open and efficient foundation language models}.
\newblock {\em arXiv preprint arXiv:2302.13971}, 2023.

\bibitem{CCpdf}
Micha{\l} Turski, Tomasz Stanis{\l}awek, Karol Kaczmarek, Pawe{\l} Dyda, and Filip Grali{\'{n}}ski.
\newblock {CCpdf: Building a High Quality Corpus for Visually Rich Documents from Web Crawl Data}.
\newblock In Gernot~A. Fink, Rajiv Jain, Koichi Kise, and Richard Zanibbi, editors, {\em Document Analysis and Recognition - ICDAR 2023}, pages 348--365, Cham, 2023. Springer Nature Switzerland.

\bibitem{unstructured_io}
{Unstructured.io Team}.
\newblock {Unstructured.io: Open-Source Pre-Processing Tools for Unstructured Data}.
\newblock https://unstructured.io, 2024.
\newblock Accessed: 2024-11-19.

\bibitem{vicuna2023}
{Vicuna}.
\newblock {Vicuna: An open-source chatbot impressing GPT-4 with 90\%* ChatGPT quality}.
\newblock \url{https://vicuna.lmsys.org/}, 2023.

\bibitem{wang2024mineruopensourcesolutionprecise}
Bin Wang, Chao Xu, Xiaomeng Zhao, Linke Ouyang, Fan Wu, Zhiyuan Zhao, Rui Xu, Kaiwen Liu, Yuan Qu, Fukai Shang, Bo~Zhang, Liqun Wei, Zhihao Sui, Wei Li, Botian Shi, Yu~Qiao, Dahua Lin, and Conghui He.
\newblock {MinerU: An Open-Source Solution for Precise Document Content Extraction}, 2024.

\bibitem{waskom2021seaborn}
Michael~L Waskom.
\newblock Seaborn: statistical data visualization.
\newblock {\em Journal of Open Source Software}, 6(60):3021, 2021.

\bibitem{wordscape}
Maurice Weber, Carlo Siebenschuh, Rory Butler, Anton Alexandrov, Valdemar Thanner, Georgios Tsolakis, Haris Jabbar, Ian Foster, Bo~Li, Rick Stevens, et~al.
\newblock {WordScape: a Pipeline to extract multilingual, visually rich Documents with Layout Annotations from Web Crawl Data}.
\newblock {\em Advances in Neural Information Processing Systems}, 36, 2024.

\bibitem{wei2024general}
Haoran Wei, Chenglong Liu, Jinyue Chen, Jia Wang, Lingyu Kong, Yanming Xu, Zheng Ge, Liang Zhao, Jianjian Sun, Yuang Peng, et~al.
\newblock {General OCR Theory: Towards OCR-2.0 via a Unified End-to-end Model}.
\newblock {\em arXiv preprint arXiv:2409.01704}, 2024.

\bibitem{wei2022chain}
Jason Wei, Xuezhi Wang, Dale Schuurmans, Maarten Bosma, Fei Xia, Ed~Chi, Quoc~V Le, Denny Zhou, et~al.
\newblock Chain-of-thought prompting elicits reasoning in large language models.
\newblock {\em Advances in neural information processing systems}, 35:24824--24837, 2022.

\bibitem{SMILES}
David Weininger.
\newblock { SMILES, a chemical language and information system. 1. Introduction to methodology and encoding rules}.
\newblock {\em Journal of Chemical Information and Computer Sciences}, 28(1):31--36, Feb 1988.

\bibitem{pandas}
{Wes McKinney and the Pandas Development Team}.
\newblock {Pandas: powerful Python data analysis toolkit, Release 1.4.4}.
\newblock \url{https://pandas.pydata.org/pandas-docs/version/1.4/pandas.pdf}, 2022.

\bibitem{xia2023structchart}
Renqiu Xia, Bo~Zhang, Haoyang Peng, Hancheng Ye, Xiangchao Yan, Peng Ye, Botian Shi, Junchi Yan, and Yu~Qiao.
\newblock {StructChart: Perception, Structuring, Reasoning for Visual Chart Understanding}.
\newblock {\em arXiv preprint arXiv:2309.11268}, 2023.

\bibitem{xia2024chartx}
Renqiu Xia, Bo~Zhang, Hancheng Ye, Xiangchao Yan, Qi~Liu, Hongbin Zhou, Zijun Chen, Min Dou, Botian Shi, Junchi Yan, et~al.
\newblock Chartx \& chartvlm: A versatile benchmark and foundation model for complicated chart reasoning.
\newblock {\em arXiv preprint arXiv:2402.12185}, 2024.

\bibitem{xu2020layoutlmv2}
Yang Xu, Yiheng Xu, Tengchao Lv, Lei Cui, Furu Wei, Guoxin Wang, Yijuan Lu, Dinei Florencio, Cha Zhang, Wanxiang Che, et~al.
\newblock Layoutlmv2: Multi-modal pre-training for visually-rich document understanding.
\newblock {\em arXiv preprint arXiv:2012.14740}, 2020.

\bibitem{xu2020layoutlm}
Yiheng Xu, Minghao Li, Lei Cui, Shaohan Huang, Furu Wei, and Ming Zhou.
\newblock Layoutlm: Pre-training of text and layout for document image understanding.
\newblock In {\em Proceedings of the 26th ACM SIGKDD international conference on knowledge discovery \& data mining}, pages 1192--1200, 2020.

\bibitem{yan2021convmath}
Zuoyu Yan, Xiaode Zhang, Liangcai Gao, Ke~Yuan, and Zhi Tang.
\newblock {ConvMath: a convolutional sequence network for mathematical expression recognition}.
\newblock In {\em 2020 25th International Conference on Pattern Recognition (ICPR)}, pages 4566--4572. IEEE, 2021.

\bibitem{ye2023ureader}
Jiabo Ye, Anwen Hu, Haiyang Xu, Qinghao Ye, Ming Yan, Guohai Xu, Chenliang Li, Junfeng Tian, Qi~Qian, Ji~Zhang, et~al.
\newblock {Ureader: Universal ocr-free visually-situated language understanding with multimodal large language model}.
\newblock {\em arXiv preprint arXiv:2310.05126}, 2023.

\bibitem{internlmxcomposer}
Pan Zhang, Xiaoyi Dong, Bin Wang, Yuhang Cao, Chao Xu, Linke Ouyang, Zhiyuan Zhao, Shuangrui Ding, Songyang Zhang, Haodong Duan, Wenwei Zhang, Hang Yan, Xinyue Zhang, Wei Li, Jingwen Li, Kai Chen, Conghui He, Xingcheng Zhang, Yu~Qiao, Dahua Lin, and Jiaqi Wang.
\newblock {InternLM-XComposer: A Vision-Language Large Model for Advanced Text-image Comprehension and Composition}.
\newblock {\em arXiv preprint arXiv:2309.15112}, 2023.

\bibitem{internlmxcomposer2_5}
Pan Zhang, Xiaoyi Dong, Yuhang Zang, Yuhang Cao, Rui Qian, Lin Chen, Qipeng Guo, Haodong Duan, Bin Wang, Linke Ouyang, Songyang Zhang, Wenwei Zhang, Yining Li, Yang Gao, Peng Sun, Xinyue Zhang, Wei Li, Jingwen Li, Wenhai Wang, Hang Yan, Conghui He, Xingcheng Zhang, Kai Chen, Jifeng Dai, Yu~Qiao, Dahua Lin, and Jiaqi Wang.
\newblock {InternLM-XComposer-2.5: A Versatile Large Vision Language Model Supporting Long-Contextual Input and Output}.
\newblock {\em arXiv preprint arXiv:2407.03320}, 2024.

\bibitem{zhang2023vision}
Qiming Zhang, Jing Zhang, Yufei Xu, and Dacheng Tao.
\newblock {Vision Transformer with Quadrangle Attention}.
\newblock {\em arXiv preprint arXiv:2303.15105}, 2023.

\bibitem{zhang2023opt}
Susan Zhang, Stephen Roller, Naman Goyal, Mikel Artetxe, Moya Chen, Shuohui Chen, Christopher Dewan, Mona Diab, Xian Li, Xi~Victoria Lin, et~al.
\newblock Opt: Open pre-trained transformer language models.
\newblock {\em URL https://arxiv. org/abs/2205.01068}, 3:19--0, 2023.

\bibitem{gte}
Xinyi Zheng, Douglas Burdick, Lucian Popa, Xu~Zhong, and Nancy Xin~Ru Wang.
\newblock {Global table extractor (gte): A framework for joint table identification and cell structure recognition using visual context}.
\newblock In {\em Proceedings of the IEEE/CVF winter conference on applications of computer vision}, pages 697--706, 2021.

\bibitem{pubtabnet}
Xu~Zhong, Elaheh ShafieiBavani, and Antonio~Jimeno Yepes.
\newblock Image-based table recognition: data, model, and evaluation.
\newblock {\em arXiv preprint arXiv:1911.10683}, 2019.

\bibitem{zhu2023minigpt}
Deyao Zhu, Jun Chen, Xiaoqian Shen, Xiang Li, and Mohamed Elhoseiny.
\newblock Minigpt-4: Enhancing vision-language understanding with advanced large language models.
\newblock {\em arXiv preprint arXiv:2304.10592}, 2023.

\end{thebibliography}
}

\clearpage \appendix \begin{figure*}[h!]
    \centering
    \includegraphics[width=\textwidth]{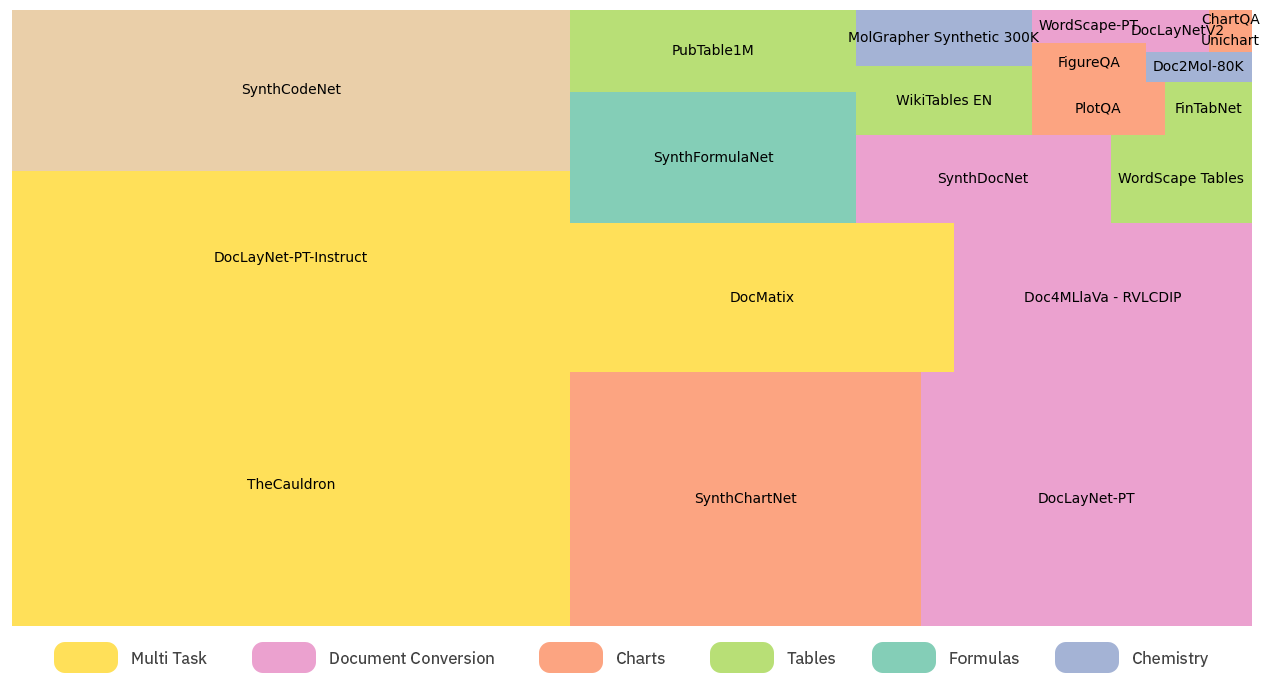}
    \caption{\textbf{Training datasets.} A treemap visualization of dataset sizes contributing to the training of SmolDocling, categorized by type. Each rectangle represents a dataset, with its area proportional to the total dataset size. Each color indicates a dataset type. \textit{``Multi-Task"} refers to datasets that include multiple different tasks.}
    \label{fig:doctags}
\end{figure*}

\begin{figure*}[htbp]
    \centering
    \includegraphics[width=0.75\textwidth]{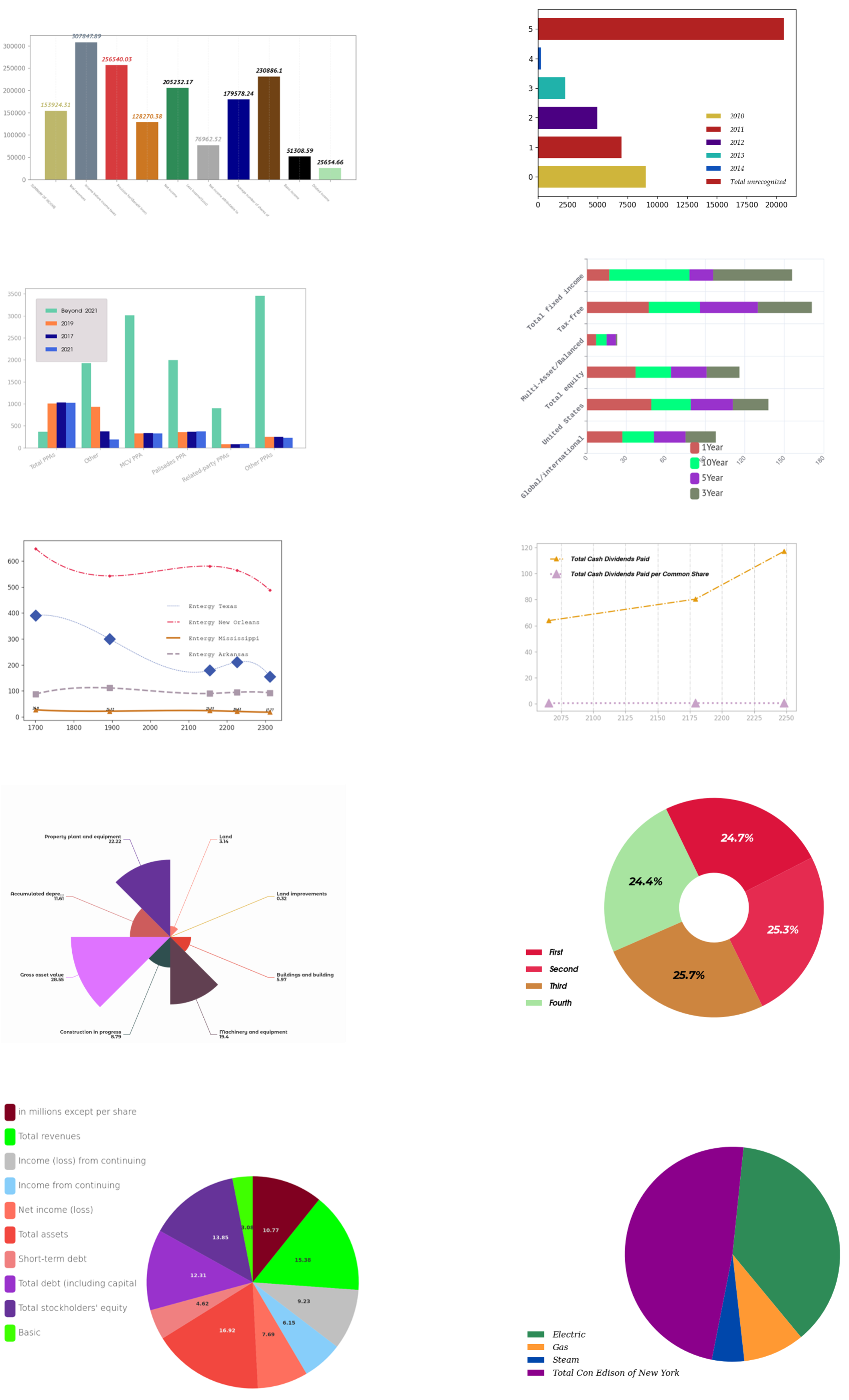}
    \caption{\textbf{Samples from SynthChartNet.}}
    \label{fig:chart_samples}
\end{figure*}

\begin{figure*}[htbp]
    \centering
    \includegraphics[width=\textwidth]{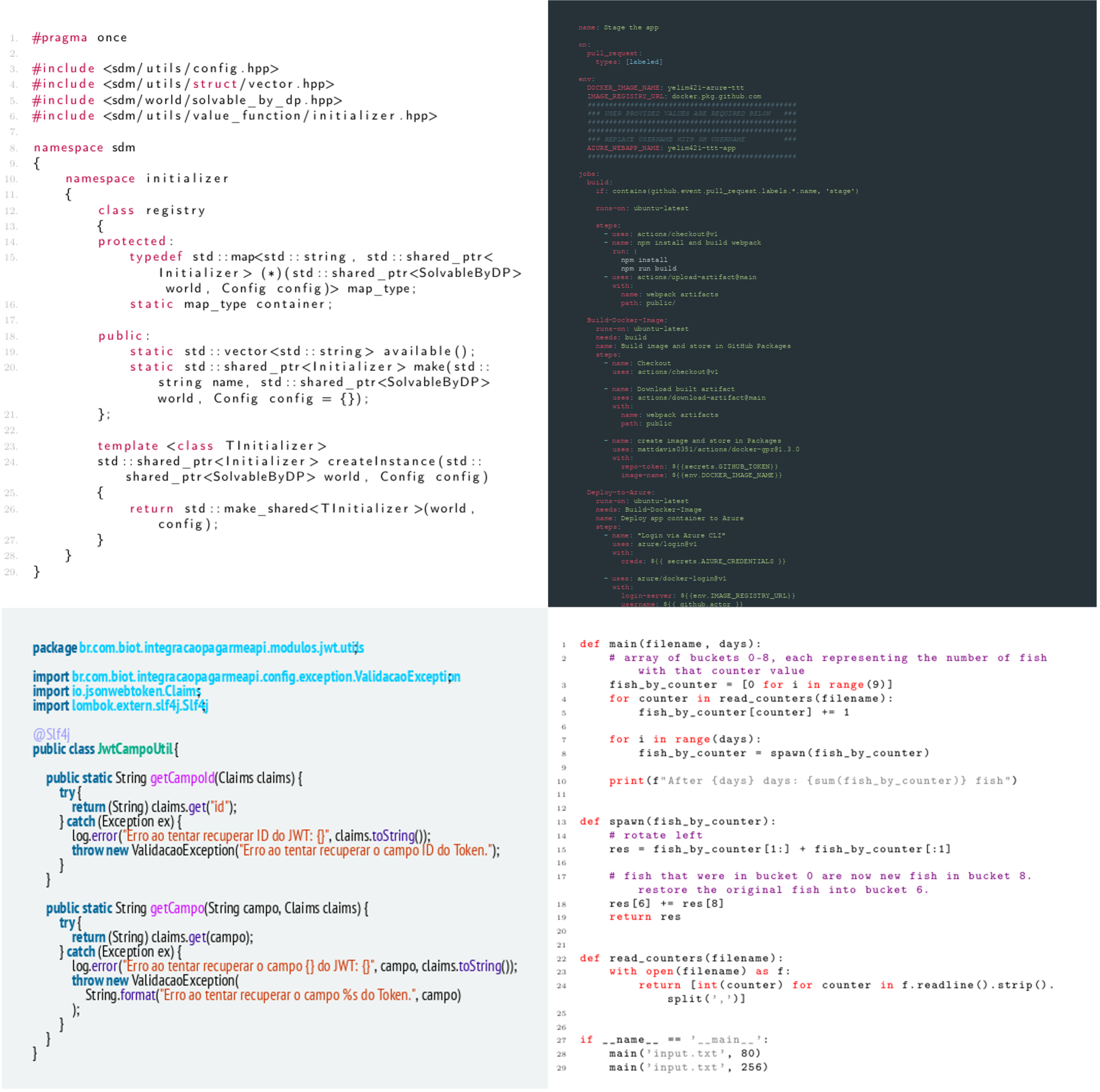}
    \caption{\textbf{Samples from SynthCodeNet.}}
    \label{fig:code_samples}
\end{figure*}

\begin{figure*}[htbp]
    \centering
    \includegraphics[width=0.8\textwidth]{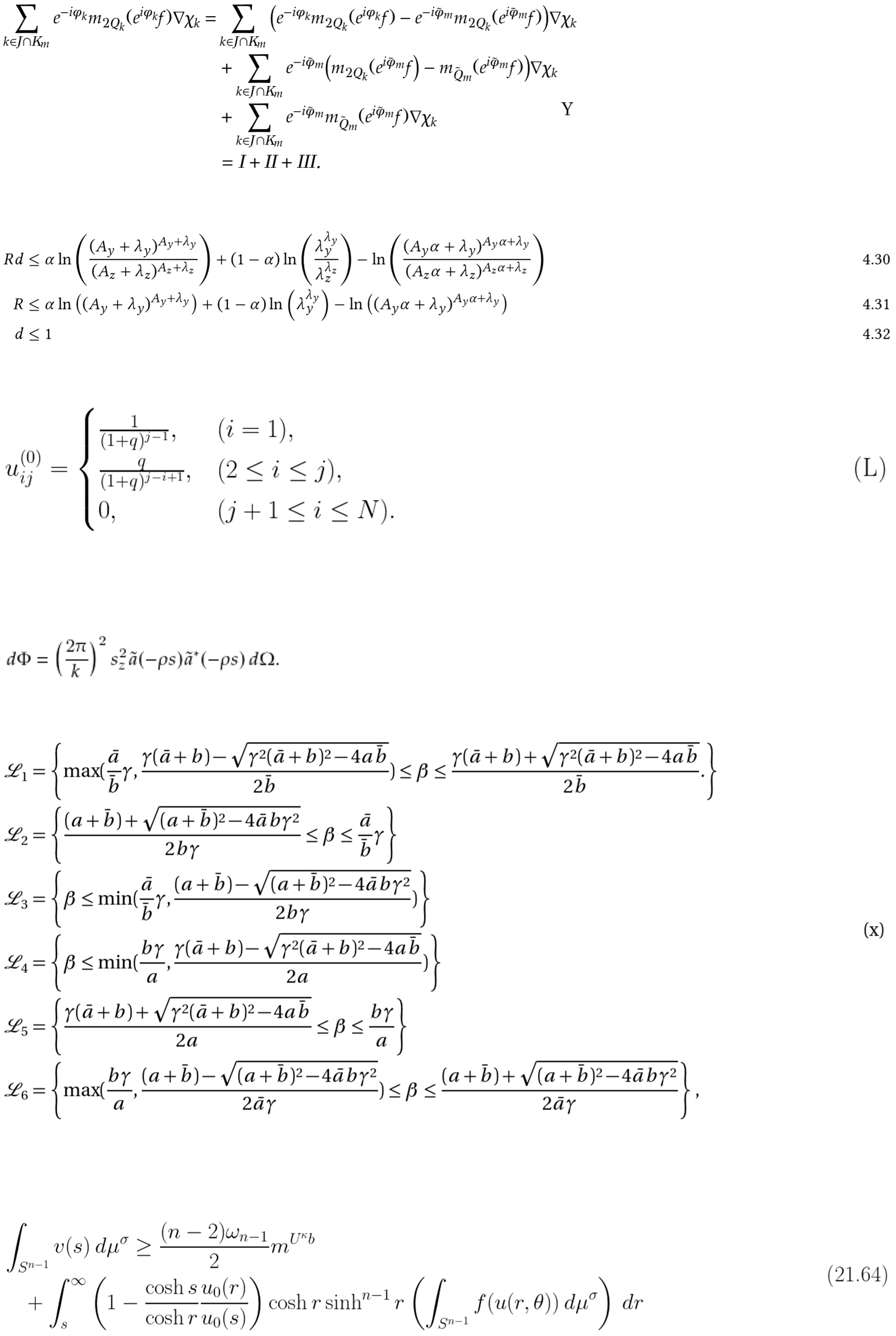}
    \caption{\textbf{Samples from SynthFormulaNet.}}
    \label{fig:formula_samples}
\end{figure*}

\section{Additional Datasets Details}
\label{sec:additional_datasets}

\subsection{SynthChartNet}

SynthChartNet includes four distinct types of charts: line, pie, bar, and stacked bar, generated using three visualization libraries: Matplotlib~\cite{matplotlib}, Seaborn~\cite{waskom2021seaborn}, and Pyecharts~\cite{pyecharts}.  Each data set was rendered through all three libraries, producing a final dataset of 2.5 million visually diverse charts. The rendering process was designed to leverage the full range of features available in these libraries, ensuring visual variety while maintaining the plausibility of the data presented. \autoref{fig:chart_samples} depicts 10 samples from SynthChartNet.

\subsection{SynthCodeNet} 
To render the code snippets of SynthCodeNet, we employed two rendering libraries: LaTeX and Pygments \cite{pygments}. LaTex, in conjunction with its listings package, enabled the generation of code renderings similar to those commonly seen in academic and technical publications, while Pygments facilitated renderings with aesthetics resembling those of IDEs. The rendering process utilized the complete set of features provided by these libraries to enhance visual diversity while preserving the plausibility of the presented data. This included randomizing various stylistic parameters such as font, font size, style, weight, line spacing, tab size, background color, line numbering (including randomizing the starting line number), and programming language-specific syntax highlighting. \autoref{fig:code_samples}
shows 4 samples from SynthCodeNet.

\subsection{SynthFormulaNet} 
The extracted formulas of SynthFormulaNet underwent a rigorous normalization procedure to ensure the model was trained only on correct and standardized code. The normalization process consisted of the following major steps:
\begin{enumerate}
    \item \textbf{Token Filtering and Removal}: Unnecessary or redundant LaTeX tokens were removed or replaced based on predefined policies. This included:
    \begin{itemize}
        \item Eliminating unwanted tokens.
        \item Replacing tokens with equivalent ones for consistency.
    \end{itemize}
    
    \item \textbf{Structural Normalization}: The equation structure was standardized by:
    \begin{itemize}
        \item Normalizing spacing between tokens.
        \item Standardizing left and right delimiters.
        \item Ensuring consistent use of tokens that require one or two braces.
    \end{itemize}
    
    \item \textbf{Token Simplification}: Redundant tokens and patterns were simplified, including:
    \begin{itemize}
        \item Collapsing consecutive redundant tokens (e.g., repeated primes, dots, or excessive spacing).
        \item Converting special constructs such as \texttt{\textbackslash Big} into \texttt{\textbackslash left} and \texttt{\textbackslash right} for consistency.
    \end{itemize}
\end{enumerate}
The extracted equations were subsequently rendered using LaTex at 120 dpi. Several aspects of the rendering procedure have been randomized to ensure the visual diversity of the dataset. Each formula was rendered using a randomly selected font from a set of 85 possible options, with variations in font size, style, weight, and line spacing. Additionally, with a certain probability, each formula was enclosed within an "align" environment, allowing for the inclusion of a randomly generated equation number on either the left or right side of the equation. The equation number itself was chosen at random from a diverse set of possibilities, including small and large numbers, integers, floating-point values, or letters. Furthermore, it could be enclosed in single or double parentheses, brackets, or curly braces and was sometimes preceded by text labels such as ``Eqn.", ``Eqn", ``Eq.", or ``Equation". \autoref{fig:formula_samples} depicts 6 samples from SynthFormulaNet.

\subsection{DocTags vocabulary}
Here we provide a reference to a complete vocabulary of \textit{DocTags} for \textit{SmolDocling}, and their meaning.

We represent tags in an XML-like style of notation. Some tags have an opening and closing part, and wrap around textual content (or other tags) for example: \texttt{<text>hello world</text>}.
Other tags are standalone and do not have a closing tag; they mark a special instruction, for example: \texttt{<page\_break>}

The complete \textit{DocTags} snippet can represent a page or multiple pages, wrapped into \texttt{<doctag>...</doctag>}. When content represents multiple pages, we use \texttt{<page\_break>} tag as a separator.

Within the scope of \texttt{<doctag>} we place high-level, tags that wrap around the textual content of certain document blocks and identify the type of the block. These tags are: \texttt{<text>}, \texttt{<caption>}, \texttt{<footnote>}, \texttt{<formula>}, \texttt{<title>}, \texttt{<page\_footer>}, \texttt{<page\_header>}, \texttt{<picture>}, \texttt{<section\_header>}, \texttt{<document\_index>}, \texttt{<code>}, \texttt{<otsl>}, \texttt{<list\_item>}, \texttt{<ordered\_list>}, \texttt{<unordered\_list>}

Each element can nest additional standalone location tags that encode its position on the page as a bounding box, represented in DocTags as \texttt{<loc\_\textit{x1}><loc\_\textit{y1}><loc\_\textit{x2}><loc\_\textit{y2}>}. Every x,y pair of coordinates belongs to a fixed grid. In our case we used integer values in the range from 0 to 500 which proportionally maps to width and height of the page.

\texttt{<otsl>} element contains table representation with following OTSL tags, that are standalone and used to describe structure of a table: \texttt{<fcel>}, \texttt{<ecel>}, \texttt{<lcel>}, \texttt{<ucel>}, \texttt{<xcel>}, \texttt{<nl>}. These tags follow rules as described in \cite{tableformer_otsl}, we additionally differentiate OTSL notion of a cell into \texttt{<fcel>} - full cell, or cell which contain content, and \texttt{<ecel>} - empty cell. We also augment tabular structure information with extra tags \texttt{<ched>}, \texttt{<rhed>}, \texttt{<srow>} that replace \texttt{<fcel>} to describe cells which belong to column and row headers of the table, as well as table section accordingly. We mark such headers when ground truth about table headers from the underlying dataset is available.

\texttt{<list\_item>} elements are placed within \texttt{<ordered\_list>} or \texttt{<unordered\_list>} and define weather it's enumerated (ordered) list or not.

\texttt{<picture>} and \texttt{<otsl>} elements by themselves can encapsulate \texttt{<caption>} tag if appropriate picture or table has it's own caption. This way we can connect extra descriptive information to illustrations and tables in the document.

\texttt{<code>} elements contain pieces of code as a content, and as such, respect tabulation and line breaks. Code elements also contains special standalone classification tag - \texttt{<\_\textit{programming-language}\_>} where \textit{programming-language} is an appropriate programming language name. Supported values (57 in total) for programming language names are: \textit{Ada}, \textit{Awk}, \textit{Bash}, \textit{bc}, \textit{C}, \textit{C\#}, \textit{C++}, \textit{CMake}, \textit{COBOL}, \textit{CSS}, \textit{Ceylon}, \textit{Clojure}, \textit{Crystal}, \textit{Cuda}, \textit{Cython}, \textit{D}, \textit{Dart}, \textit{dc}, \textit{Dockerfile}, \textit{Elixir}, \textit{Erlang}, \textit{FORTRAN}, \textit{Forth}, \textit{Go}, \textit{HTML}, \textit{Haskell}, \textit{Haxe}, \textit{Java}, \textit{JavaScript}, \textit{Julia}, \textit{Kotlin}, \textit{Lisp}, \textit{Lua}, \textit{Matlab}, \textit{MoonScript}, \textit{Nim}, \textit{OCaml}, \textit{ObjectiveC}, \textit{Octave}, \textit{PHP}, \textit{Pascal}, \textit{Perl}, \textit{Prolog}, \textit{Python}, \textit{Racket}, \textit{Ruby}, \textit{Rust}, \textit{SML}, \textit{SQL}, \textit{Scala}, \textit{Scheme}, \textit{Swift}, \textit{TypeScript}, \textit{unknown}, \textit{VisualBasic}, \textit{XML}, \textit{YAML}

Within \texttt{<picture>} elements we include picture classification with extra standalone tags \texttt{<\textit{image\_class>}}. We classify images from our datasets into following categories: \textit{natural\_image}, \textit{pie\_chart}, \textit{bar\_chart}, \textit{line\_chart}, \textit{flow\_chart}, \textit{scatter\_chart}, \textit{heatmap}, \textit{remote\_sensing}, \textit{chemistry\_molecular\_structure}, \textit{chemistry\_markush\_structure}, \textit{icon}, \textit{logo}, \textit{signature}, \textit{stamp}, \textit{qr\_code}, \textit{bar\_code}, \textit{screenshot}, \textit{map}, \textit{stratigraphic\_chart}, \textit{cad\_drawing}, \textit{electrical\_diagram}

\section{Additional Details about the SmolDocling}

\FloatBarrier

\begin{table*}[h!]
\centering
\label{tab:supported_instructions}
\begin{tabularx}{\linewidth}{>{\raggedright\arraybackslash}m{2cm} X}
\toprule
\textbf{Instruction} & \textbf{Description} \\[5pt]
\midrule
Full conversion & Convert this page to docling. \\[5pt]
\midrule
Chart & Convert chart to table (e.g., \texttt{<chart>}). \\[5pt]
\midrule
Formula & Convert formula to LaTeX (e.g., \texttt{<formula>}). \\[5pt]
\midrule
Code & Convert code to text (e.g., \texttt{<code>}). \\[5pt]
\midrule
Table & Convert table to OTSL (e.g., \texttt{<otsl>}). \\[5pt]
\midrule
No-Code Actions/Pipelines & 
\begin{tabular}[t]{@{}l@{}}
OCR the text in a specific location: \texttt{<loc\_155><loc\_233><loc\_206><loc\_237>}\\[5pt]
\specialrule{0.3pt}{0pt}{6pt}
Identify element at: \texttt{<loc\_247><loc\_482><loc\_252><loc\_486>}\\[5pt]
\specialrule{0.3pt}{0pt}{6pt}
Find all `text' elements on the page, retrieve all section headers.\\[5pt]
\specialrule{0.3pt}{0pt}{6pt}
Detect footer elements on the page.\\[5pt]
\end{tabular} \\ 
\bottomrule
\end{tabularx}
\caption{\textbf{Supported Conversion Instructions.}}
\end{table*}

\clearpage
\section{Additional Qualitative Analysis}

\subsection{Molecule Image Recognition Experiment}

Extracting molecule images from documents has potential to accelerate research in chemistry and materials discovery~\cite{Pyzer-Knapp2025}.
Some recent document understanding models, such as GOT-OCR 2.0~\cite{wei2024general} and Qwen2.5-VL~\cite{qwen2_5vl_tech_report}, claim to perform molecule image recognition in documents.
Following this direction, we experimented with training SmolDocling for molecule image recognition on full document pages and cropped images. In order to predict molecule structures (graphs) with SmolDocling, we use the molecule string identifier named SMILES~\cite{SMILES}.

\noindent\textbf{Datasets.} For training on cropped images, we use the MolGrapher-Synthetic-300K dataset~\cite{Morin_2023_ICCV}. The dataset contains 300K synthetic samples. It is created using real chemical-structures retrieved from the database PubChem~\cite{10.1093/nar/gky1033} which are then synthetically drawn using the library RDKit~\cite{RDKit}. 
For training of full pages, we use Doc2Mol-80K, a subset of the PatCID dataset~\cite{Morin2024}. Our subset contains 80 000 real patent documents published in the United-States between 2022 and 2024.

\begin{figure*}[t]
\vspace{0mm}
\centering%
\includegraphics[trim={0cm 5cm 0cm 5cm}, clip, width=0.97\textwidth]{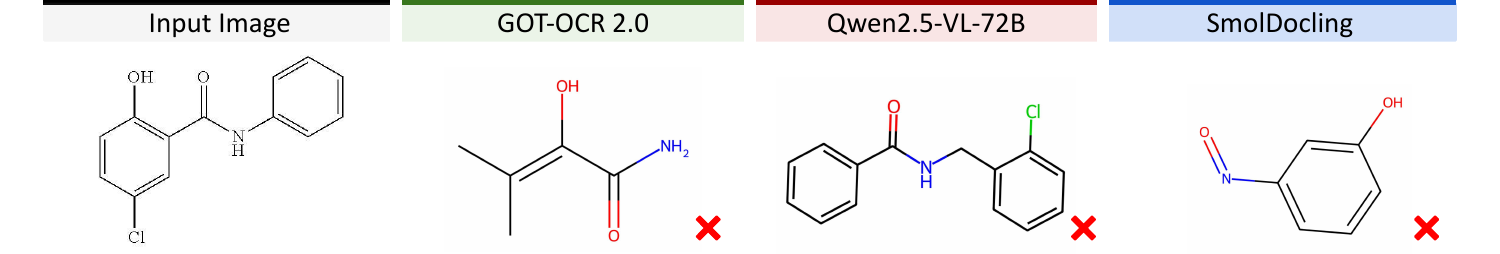}\vspace{0mm}
\caption{\textbf{Qualitative comparison for molecule image recognition.} Predictions are shown for GOT-OCR 2.0, Qwen2.5-VL-72B and SmolDocling on a simple molecule image.}
\label{fig:molecule-comparison}\vspace{-3mm}
\end{figure*}

\noindent\textbf{Evaluation.} 
\autoref{fig:molecule-comparison} shows an example of predicted molecules for a simple molecule image. 
While some models capture high-level molecular features, they consistently fail to reconstruct its detailed structure. Notably, this molecule is trivial for specialized models such as MolGrapher~\cite{Morin_2023_ICCV}, MolScribe~\cite{Qian2023} or DECIMER~\cite{Rajan2023}, which accurately recognize its structure.
Reducing the gap between general document understanding models and specialized models may require incorporating a specialized vocabulary for SMILES tokens, explicitly encoding atoms and bonds, and extending the SMILES sequence with localization tokens for each atom.

\subsection{Layout analysis samples}

To illustrate and compare the visual grounding capabilities of SmolDocling, a set of sample predictions with SmolDocling and Qwen2.5-VL from DocLayNet~\cite{doclaynet} is shown in Table~\ref{tab:layout_comparison}. Note that the element location results are independent from the correct reproduction of document content and structure. 

\begin{longtable}{|c|ccc|c|}
    \caption{Visualizations of layout output from SmolDocling and QwenVL-2.5 compared to the DocLayNet ground truth. Examples are chosen to be representative of different layout styles and features. The prediction results however do not represent a generalizable measure of the model's performance on inputs with similar features. (1) Multi-column pages are handled by SmolDocling and Qwen2.5-VL, with some recall errors in the latter. (2) A manual page with terminal output shows poor bounding box recall on SmolDocling and label confusion in Qwen2.5-VL. (3) Lists with nesting are handled well in SmolDocling but confuse Qwen2.5-VL. (4) Both SmolDocling and Qwen2.5-VL reconstruct equations well, however with different annotation conventions on including or excluding the equation index. (5) On a portrait page with colorful elements and gradient background, SmolDocling creates less accurate bounding-boxes and Qwen2.5-VL suffers low recall. (6) On a report page with tables and diagrams, SmolDocling output exhibits some repetition loop, fabricating non-existent text cells (bottom left), while Qwen2.5-VL is confused between tables and pictures.}
    \label{tab:layout_comparison} \\
    \hline
    \# & \textbf{Ground-Truth} & \textbf{SmolDocling} & \textbf{Qwen2.5-VL} & \textbf{ID} \\
    \hline
    \endhead
    
    \hline
    \multicolumn{5}{|r|}{{Continued on next page}} \\
    \hline
    \endfoot
    
    \hline
    \endlastfoot
    
    1 & \includegraphics[width=0.31\columnwidth]{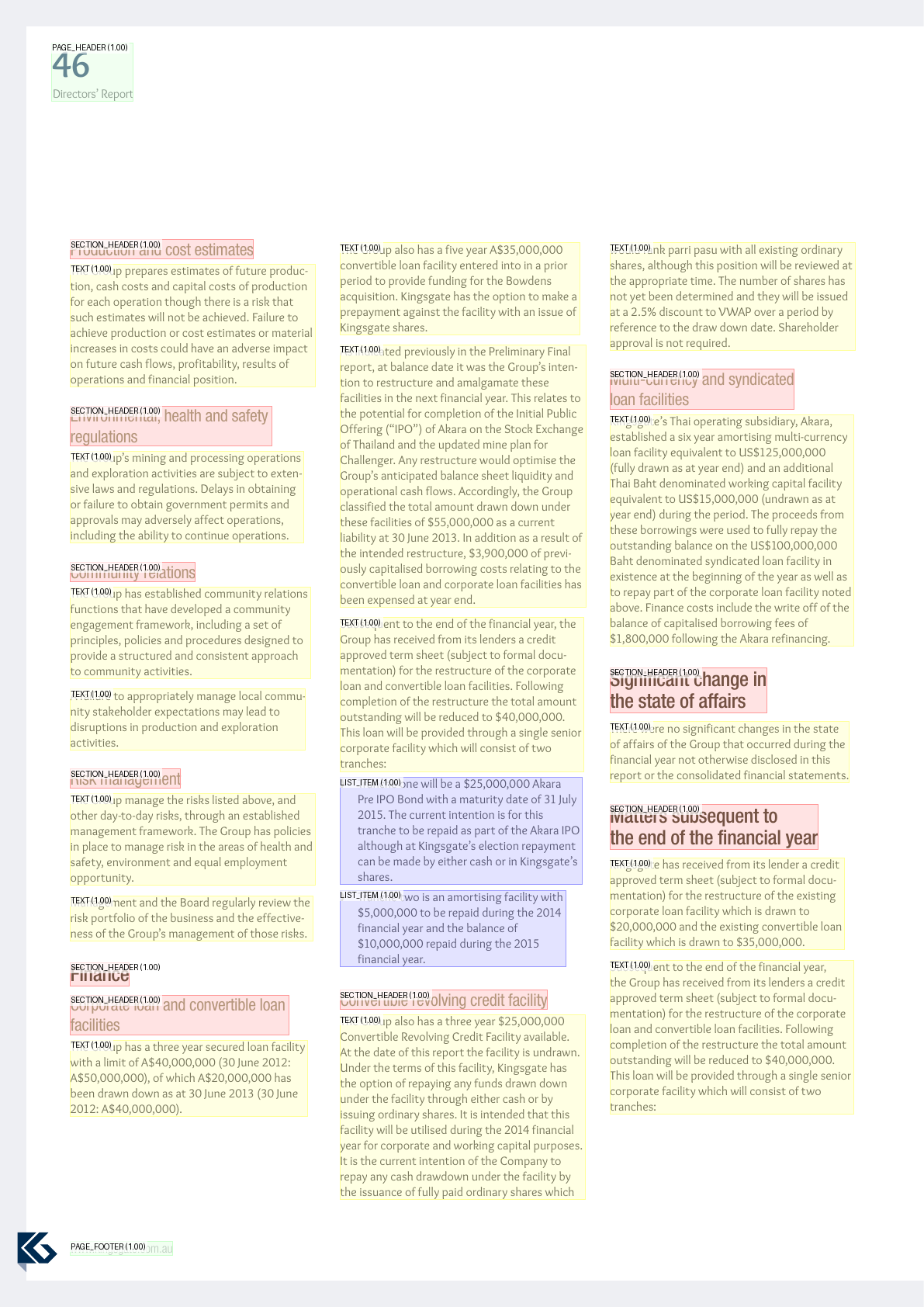} & 
    \includegraphics[width=0.31\columnwidth]{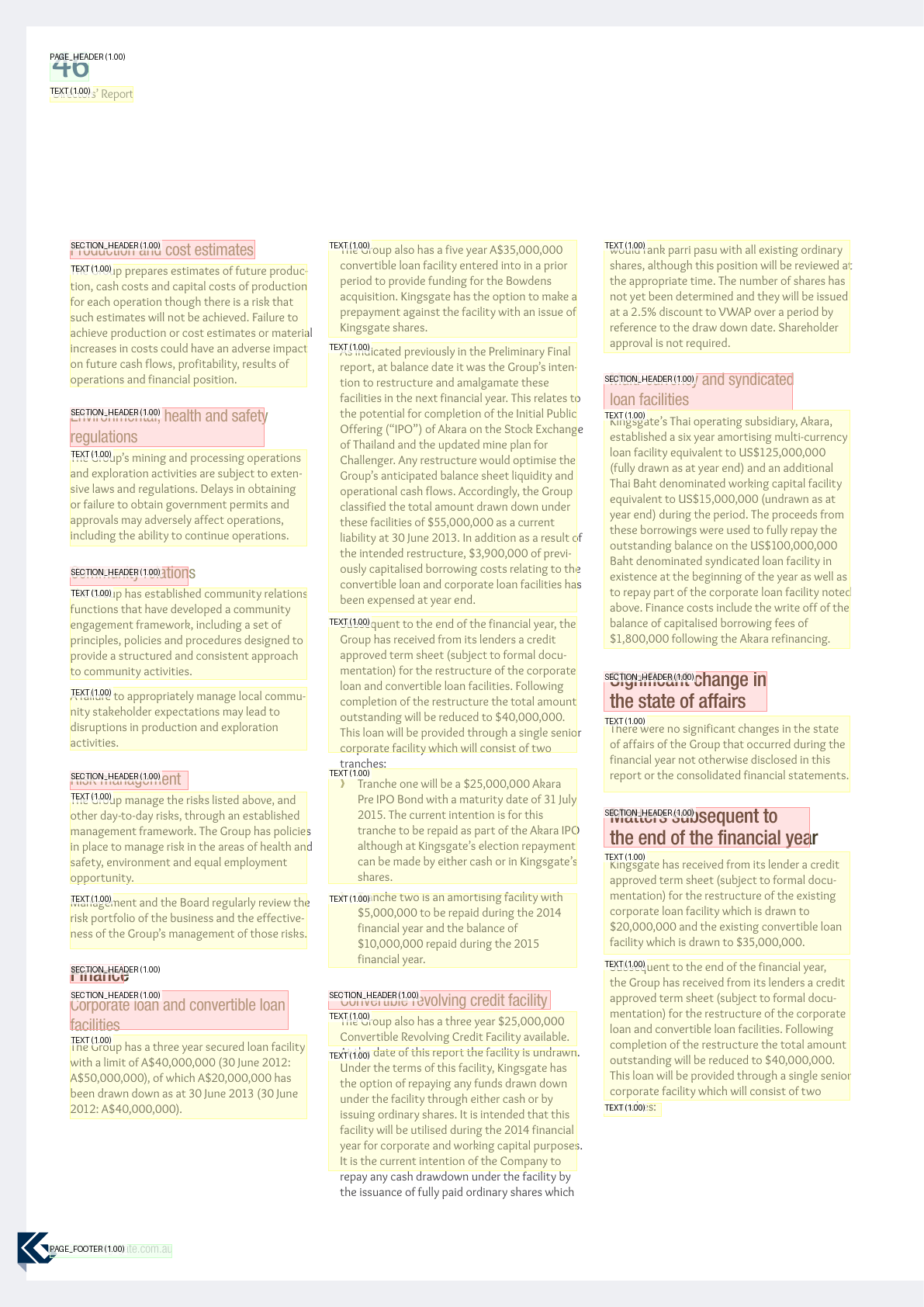} & 
    \includegraphics[width=0.31\columnwidth]{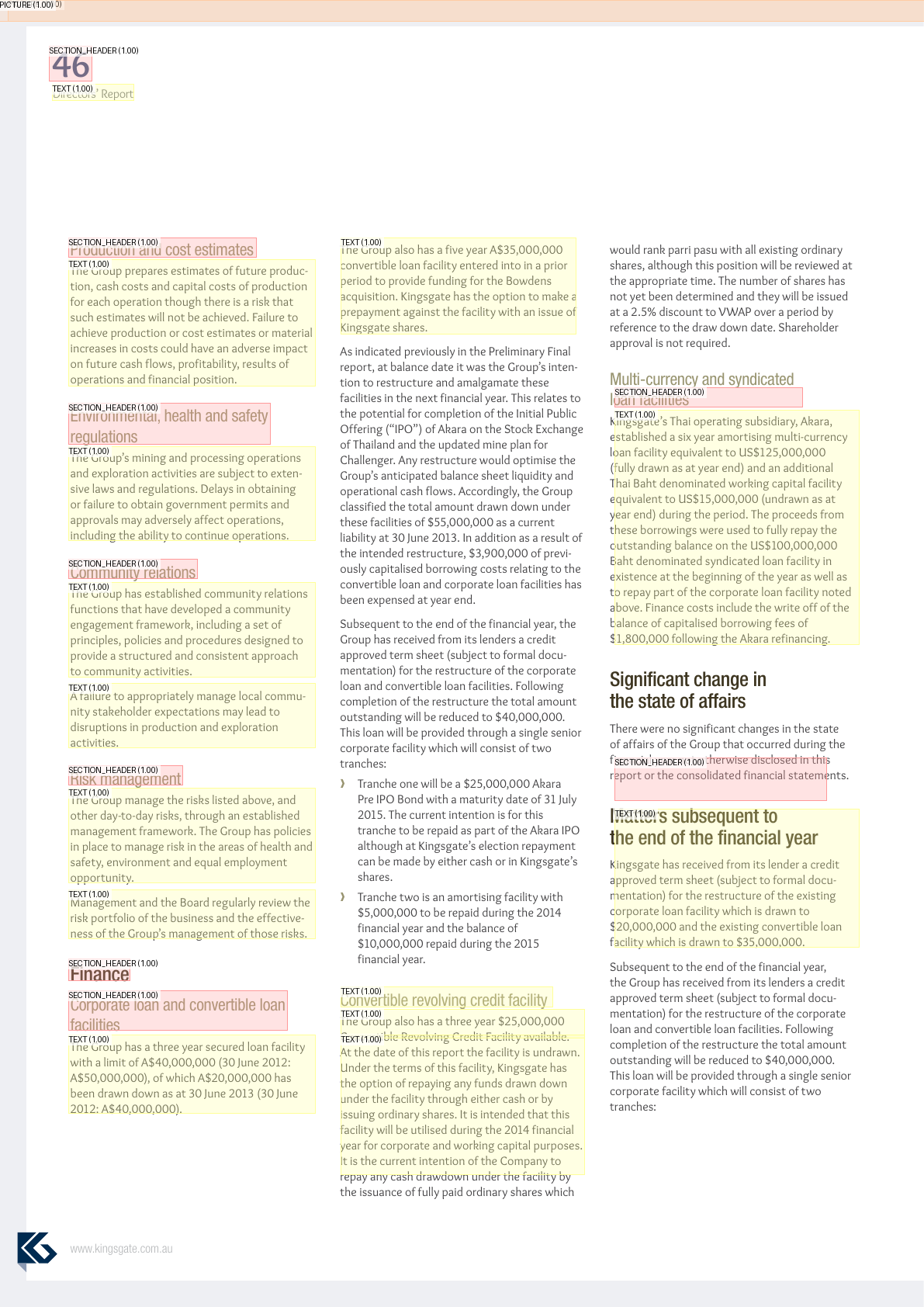} & 
    \rotatebox{90}{\tiny\ttfamily 0b49b1c3d5217411a7c9b92d960d5f61} \\
    \hline
    2 & \includegraphics[width=0.31\columnwidth]{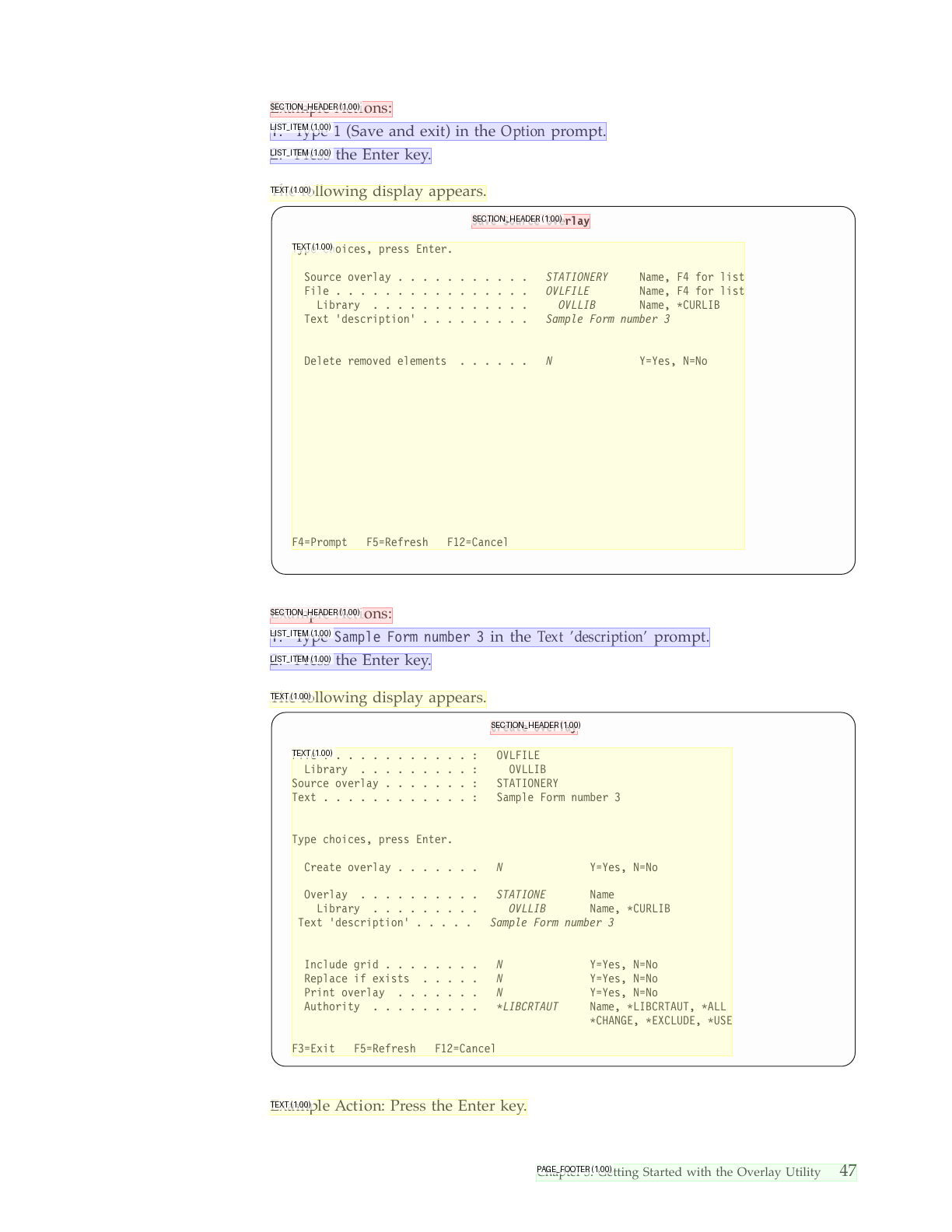} & 
    \includegraphics[width=0.31\columnwidth]{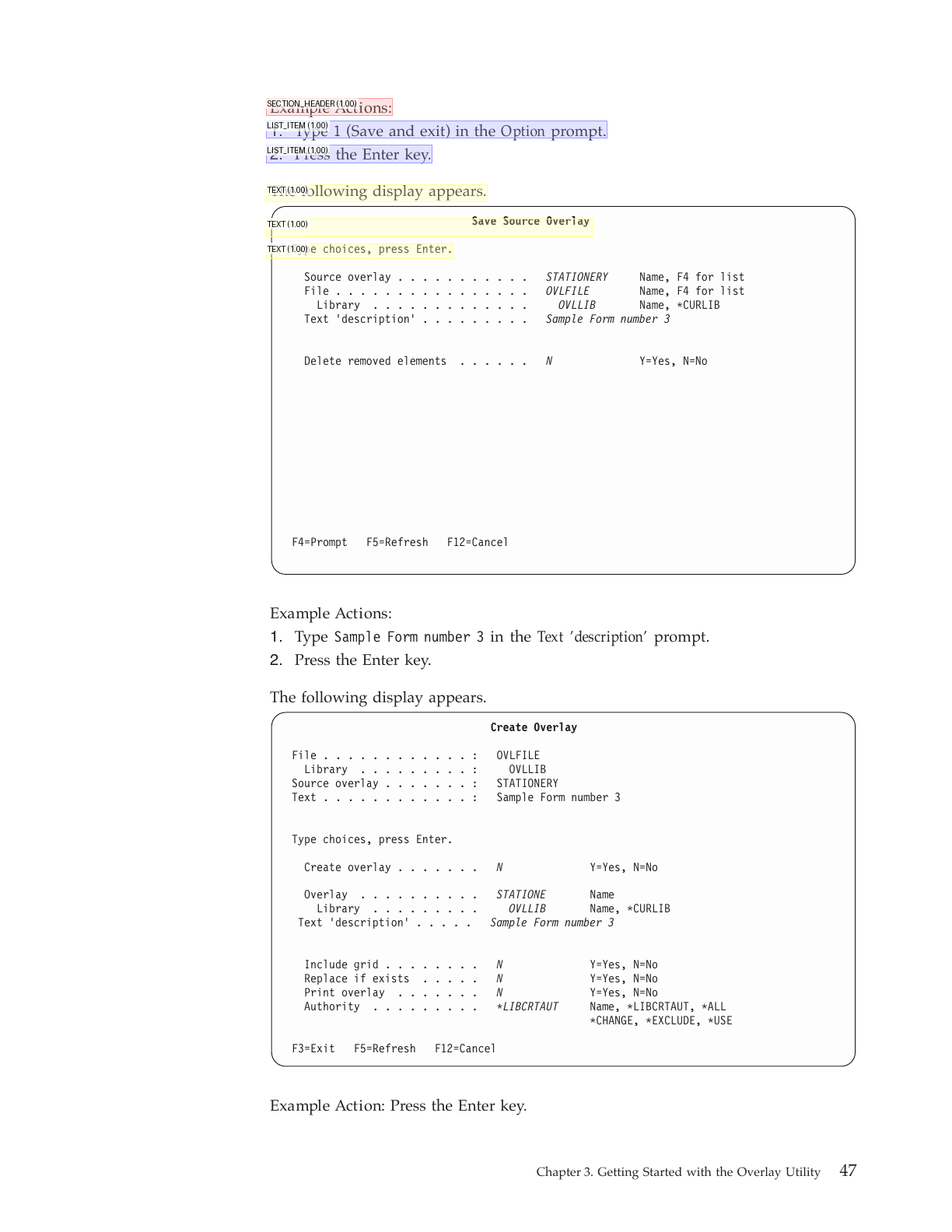} & 
    \includegraphics[width=0.31\columnwidth]{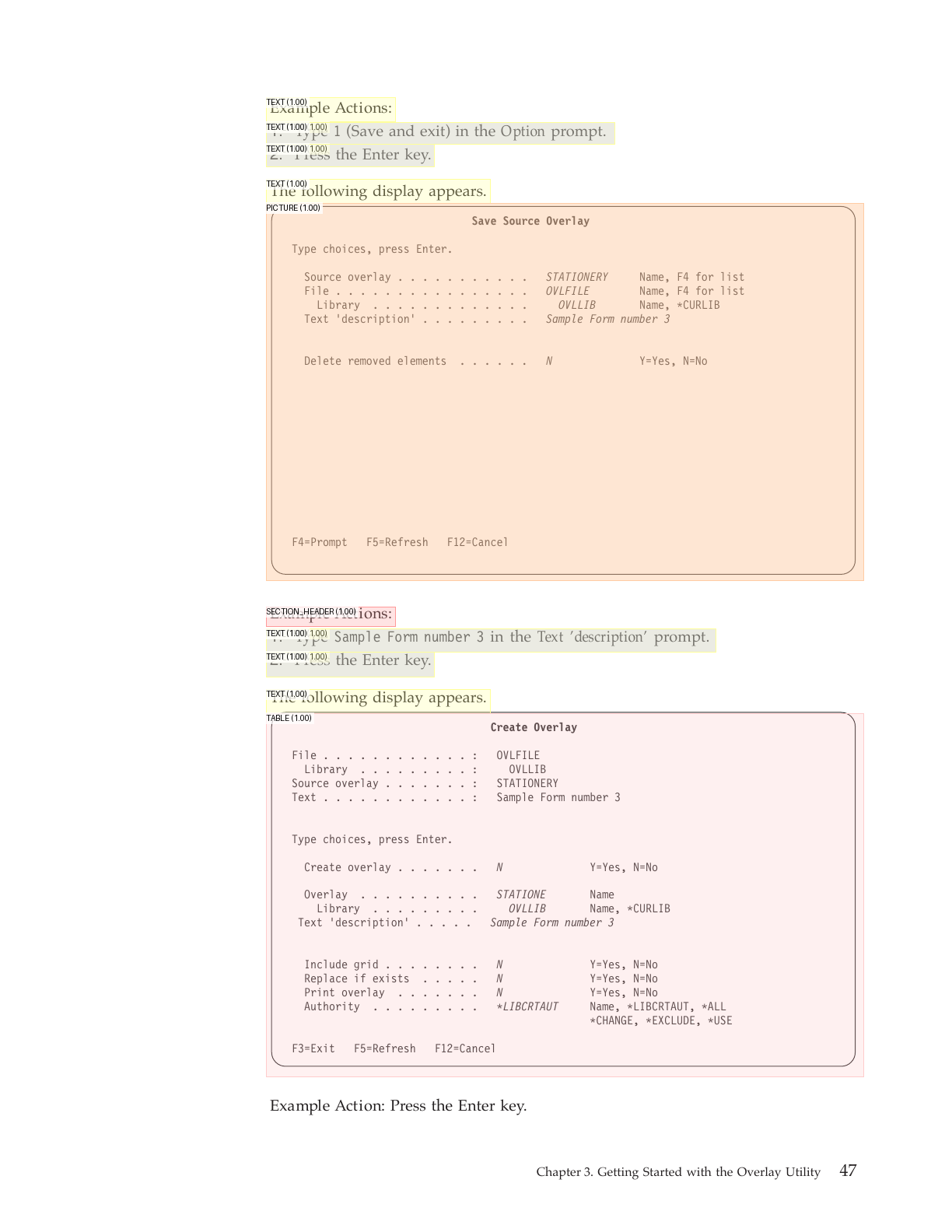} & 
    \rotatebox{90}{\tiny\ttfamily 0aafe8d4b66ac7ab81e9456327fbdb01} \\

    \hline
    3 & \includegraphics[width=0.31\columnwidth]{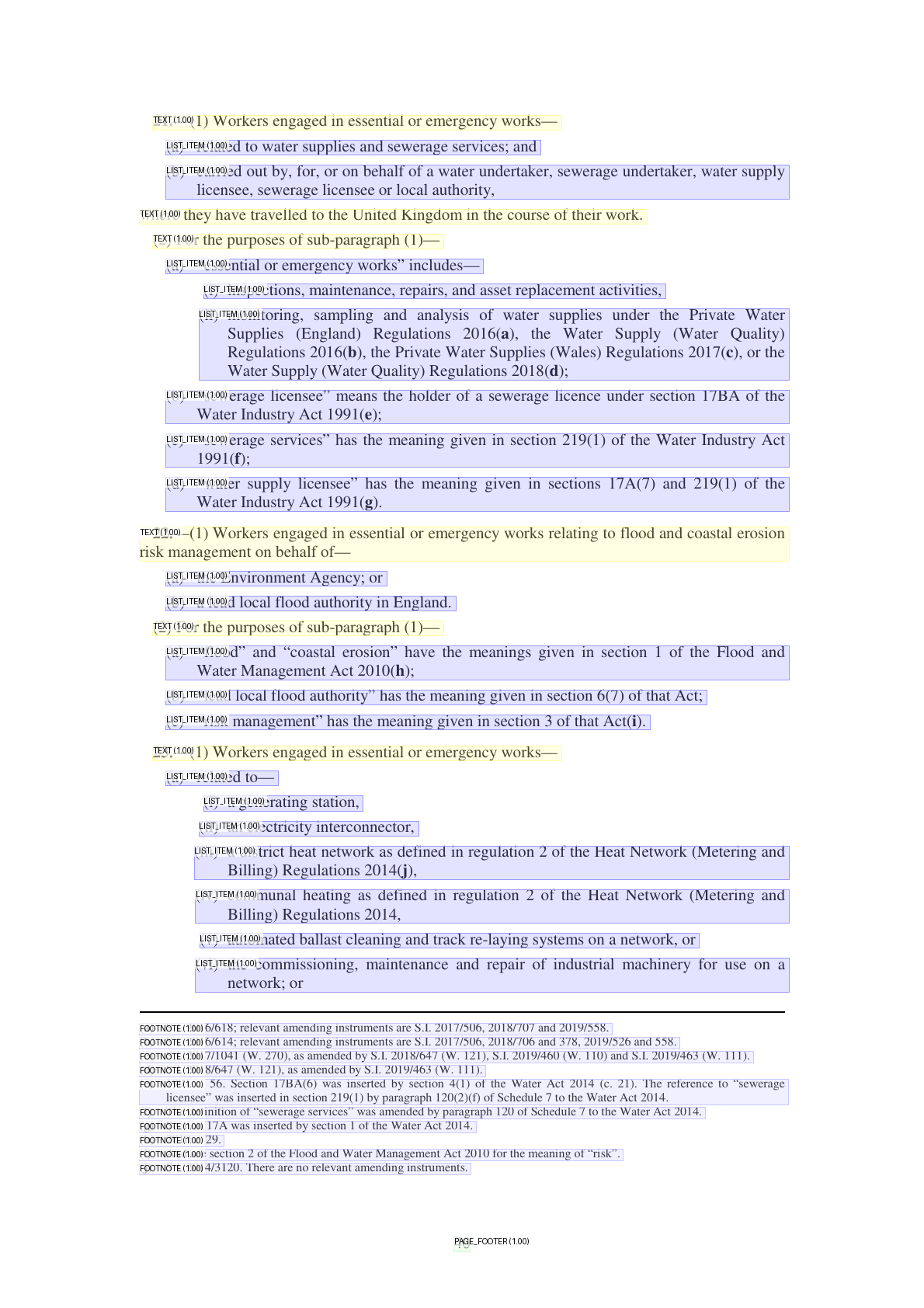} & 
    \includegraphics[width=0.31\columnwidth]{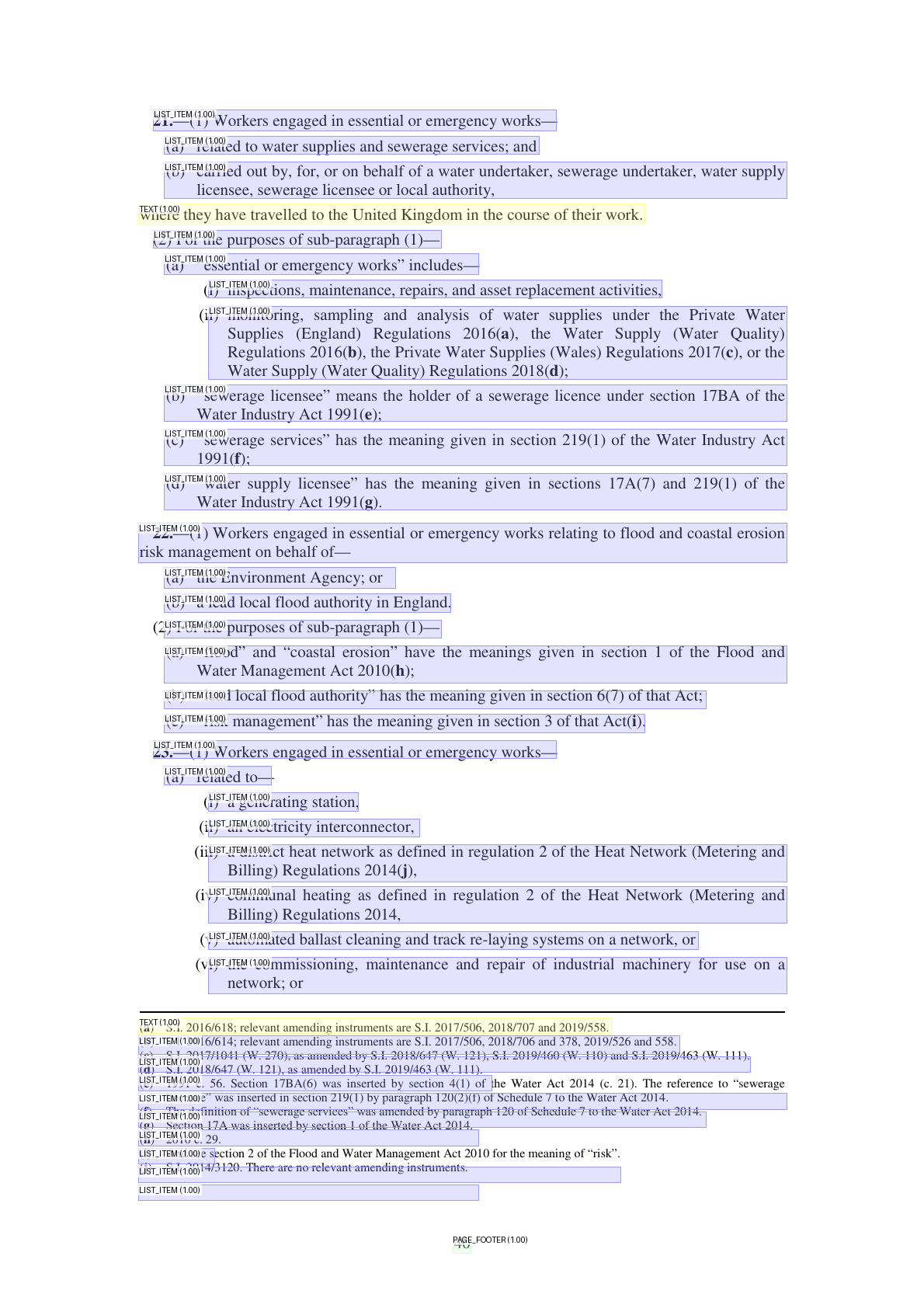} & 
    \includegraphics[width=0.31\columnwidth]{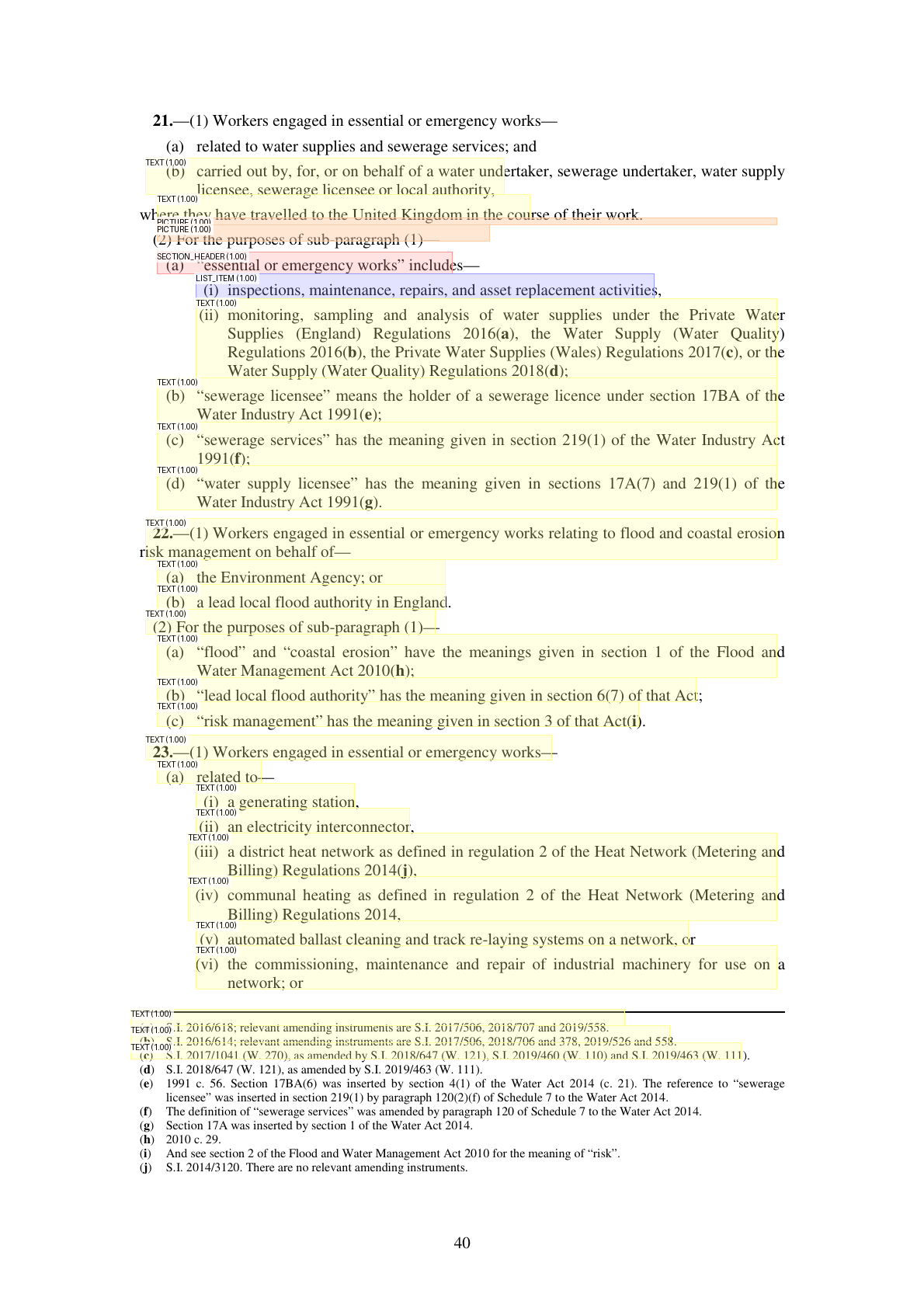} & 
    \rotatebox{90}{\tiny\ttfamily 1bff94fbe34e976d53eabc5b8b43f355} \\
    \hline
    4 & \includegraphics[width=0.31\columnwidth]{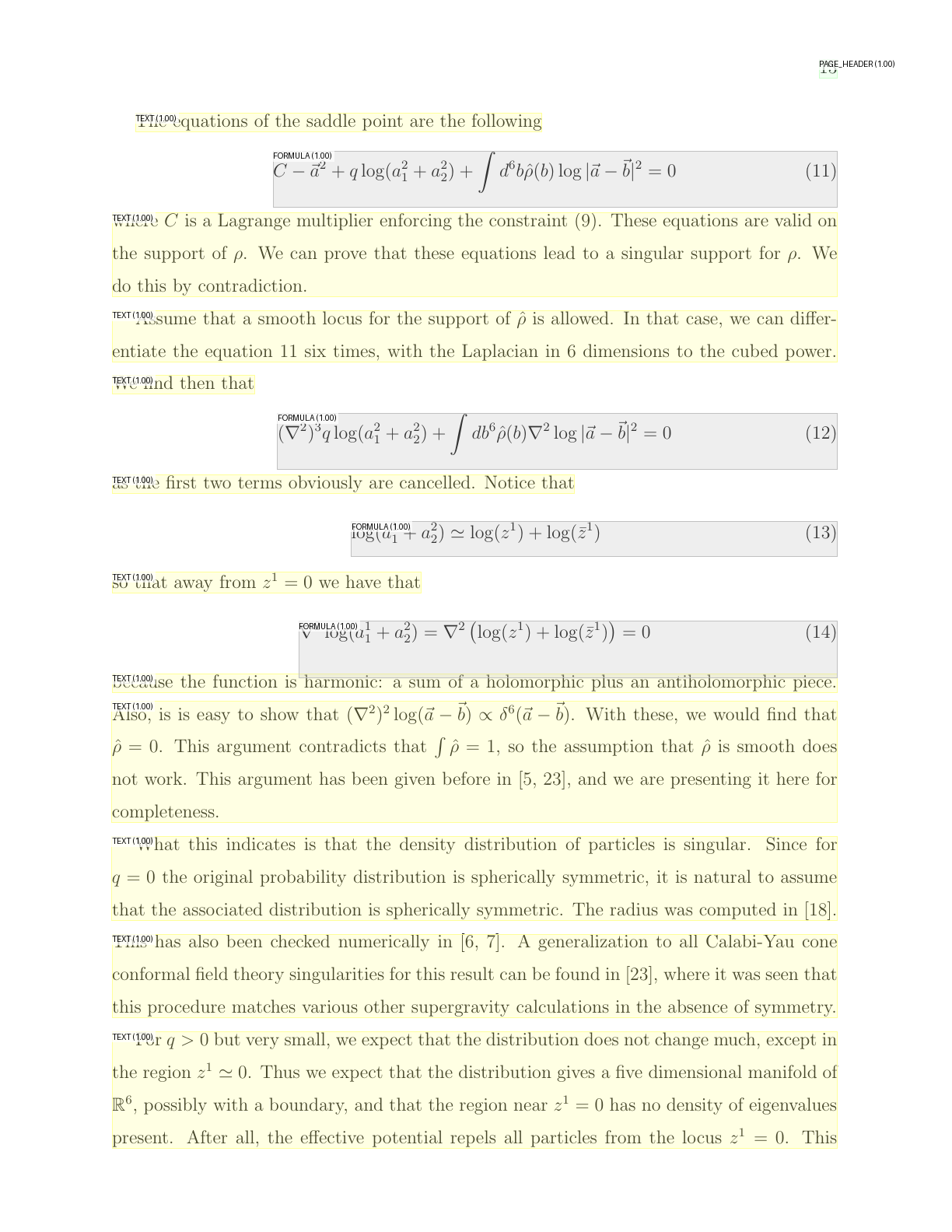} & 
    \includegraphics[width=0.31\columnwidth]{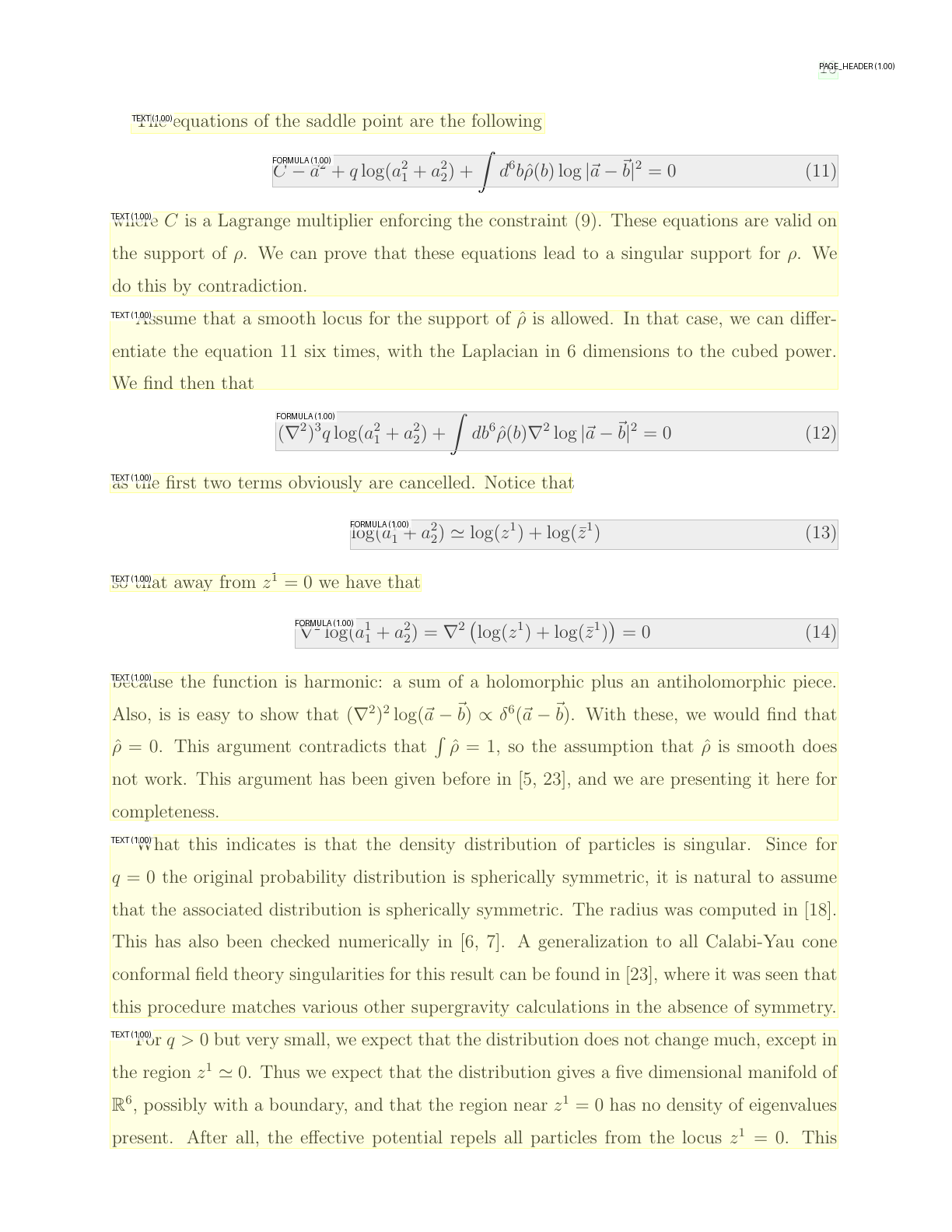} & 
    \includegraphics[width=0.31\columnwidth]{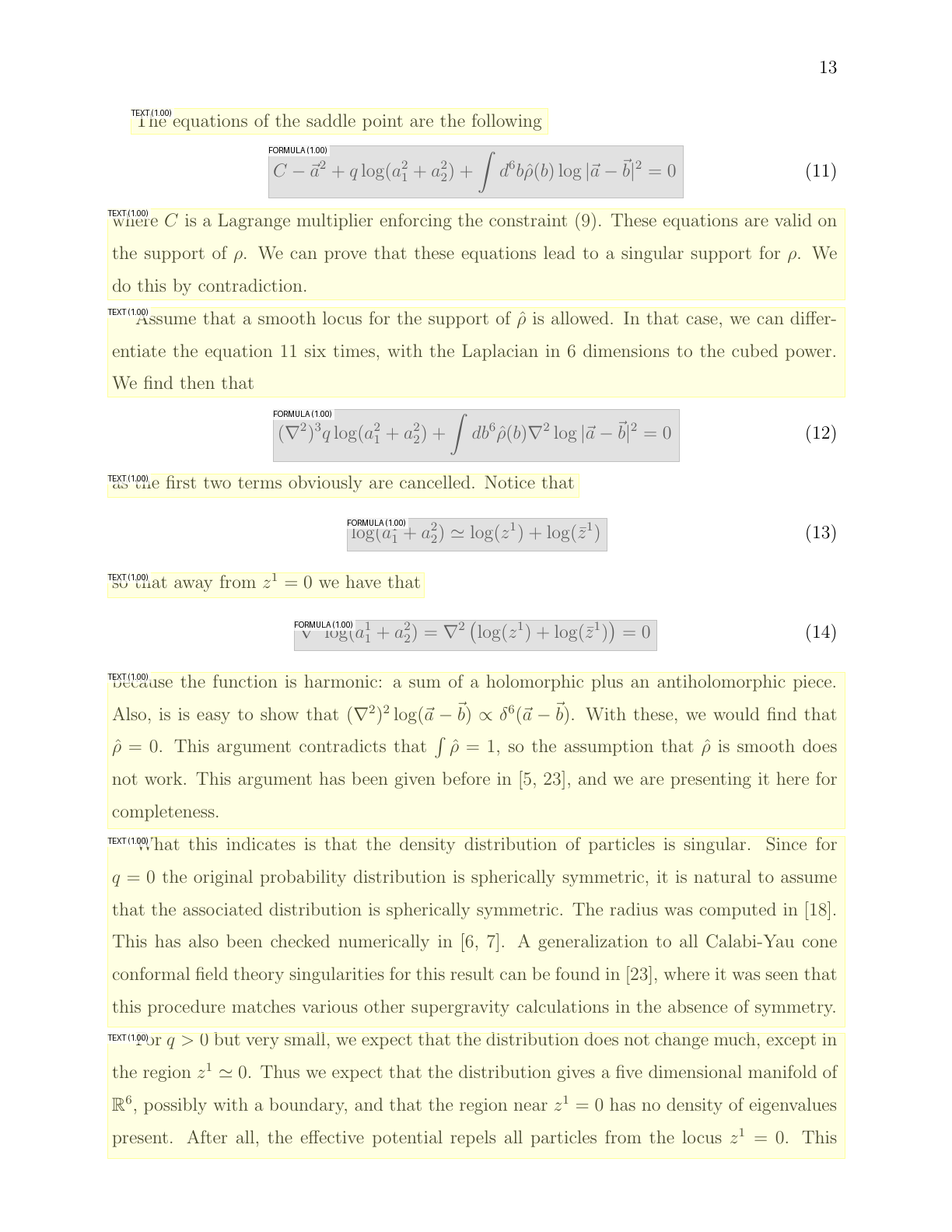} & 
    \rotatebox{90}{\tiny\ttfamily 2ab06e940ef25ff7b2e832ab8176de15} \\
    \hline
    5 & \includegraphics[width=0.31\columnwidth]{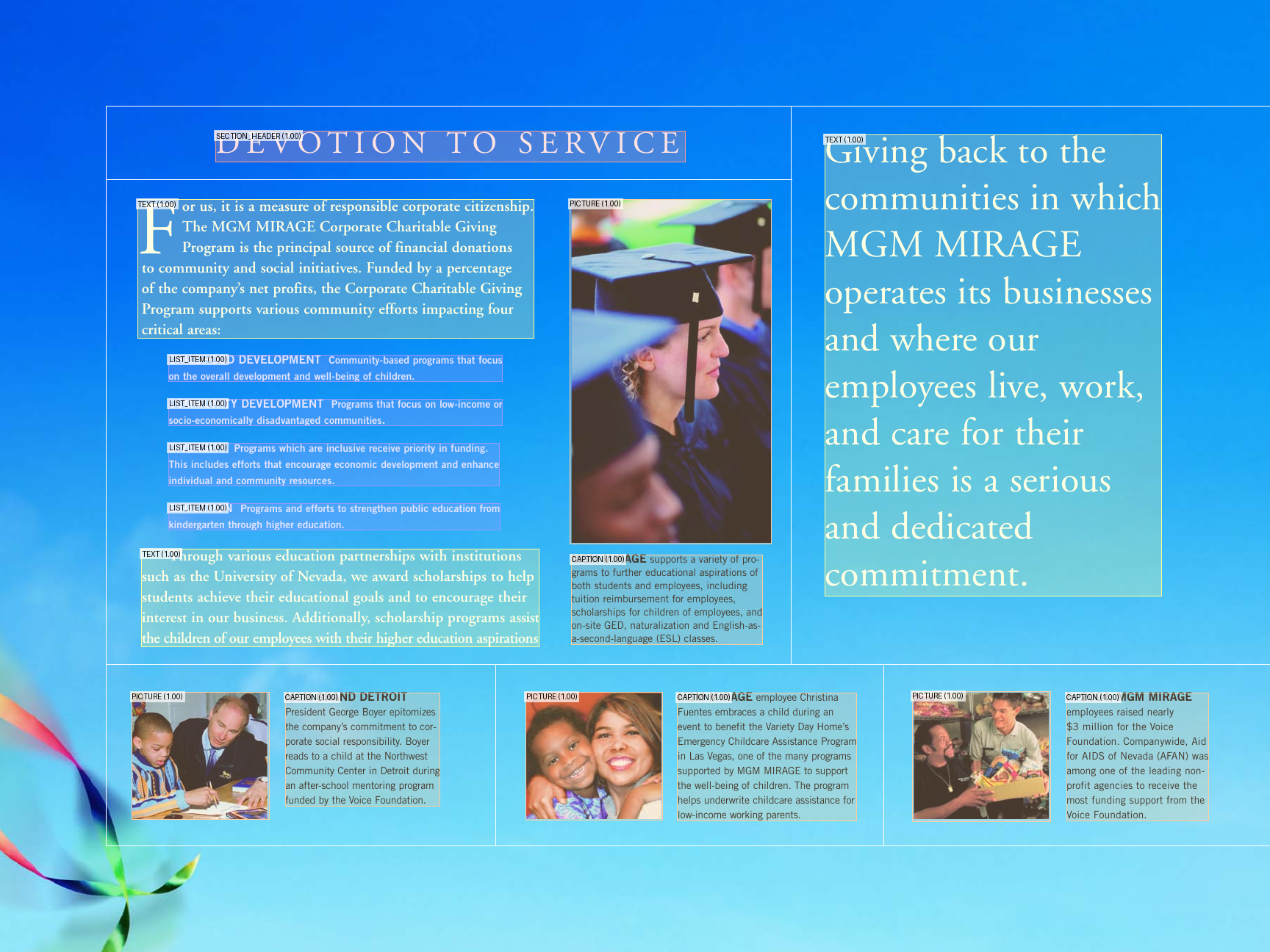} & 
    \includegraphics[width=0.31\columnwidth]{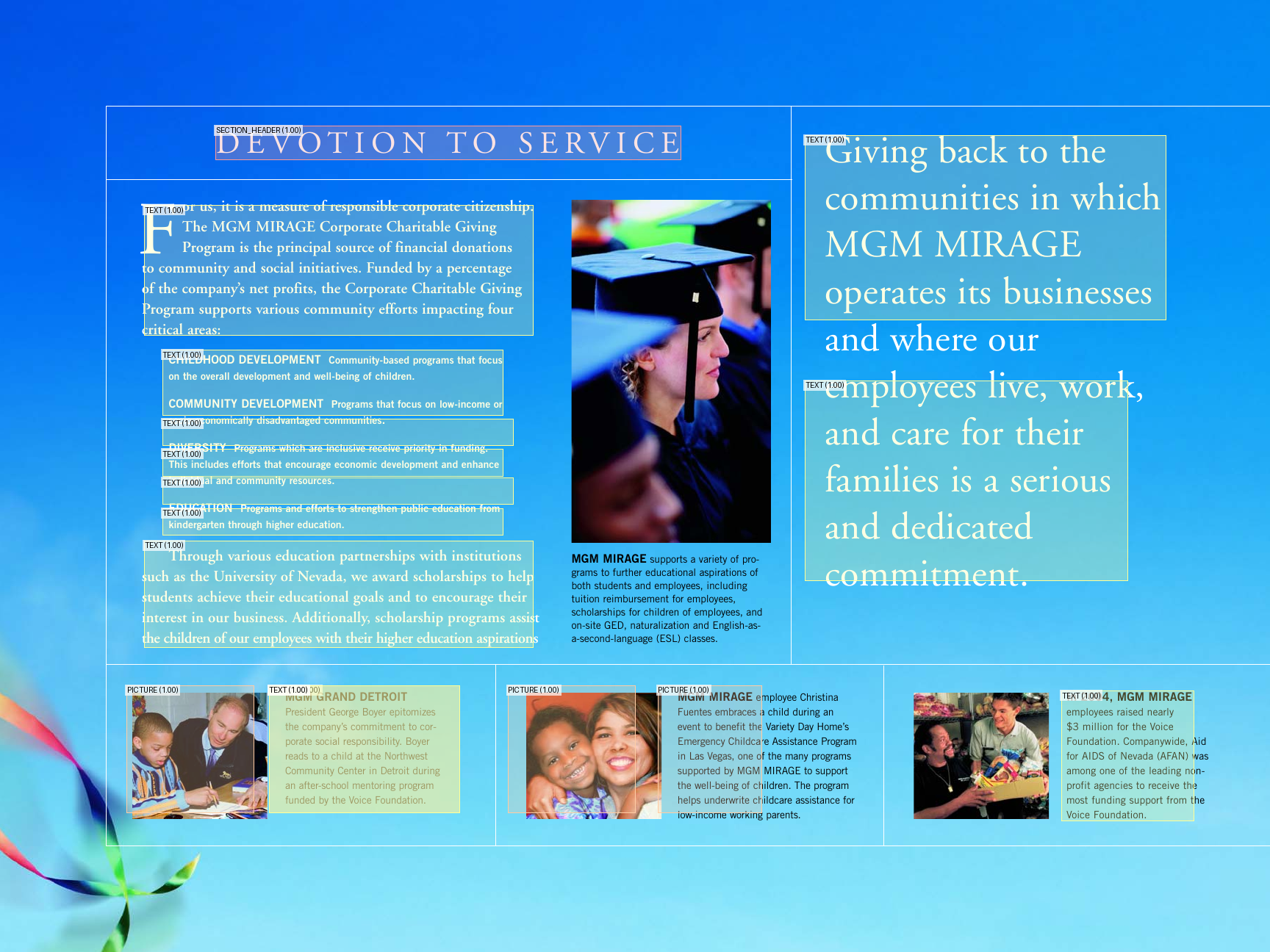} & 
    \includegraphics[width=0.31\columnwidth]{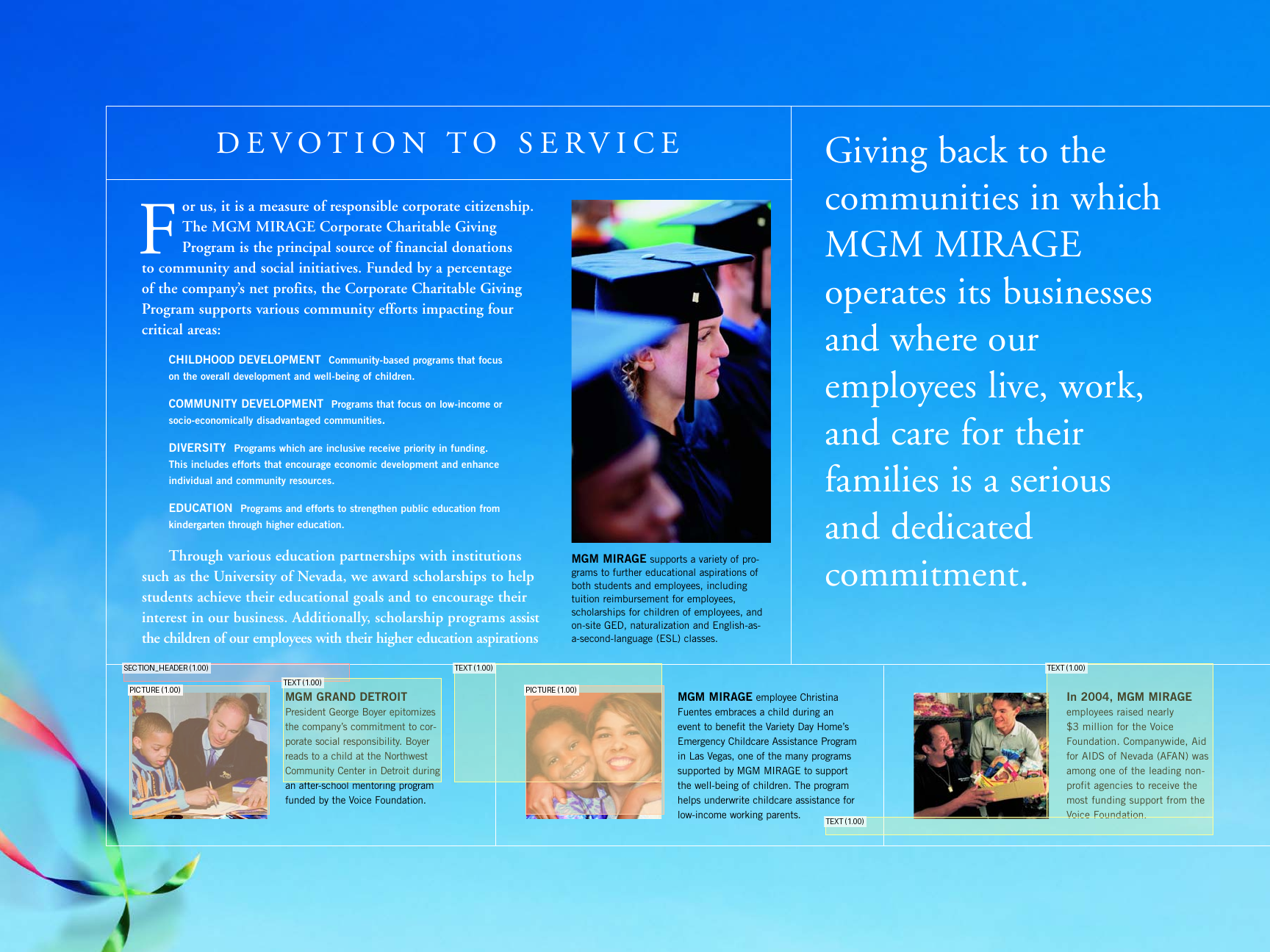} & 
    \rotatebox{90}{\tiny\ttfamily ebd1b976af12cd1d16d58939049920df} \\
    \hline
    6 & \includegraphics[width=0.31\columnwidth]{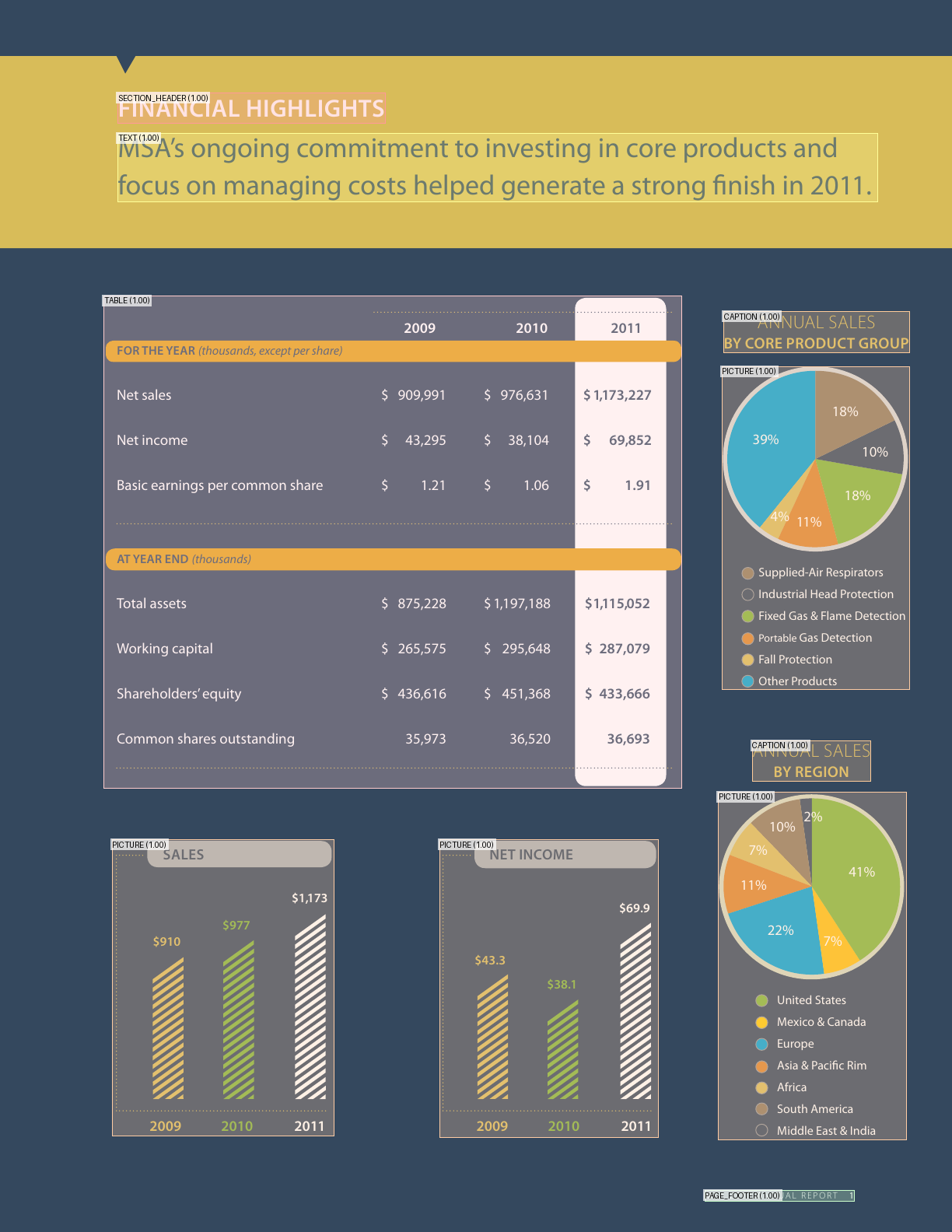} & 
    \includegraphics[width=0.31\columnwidth]{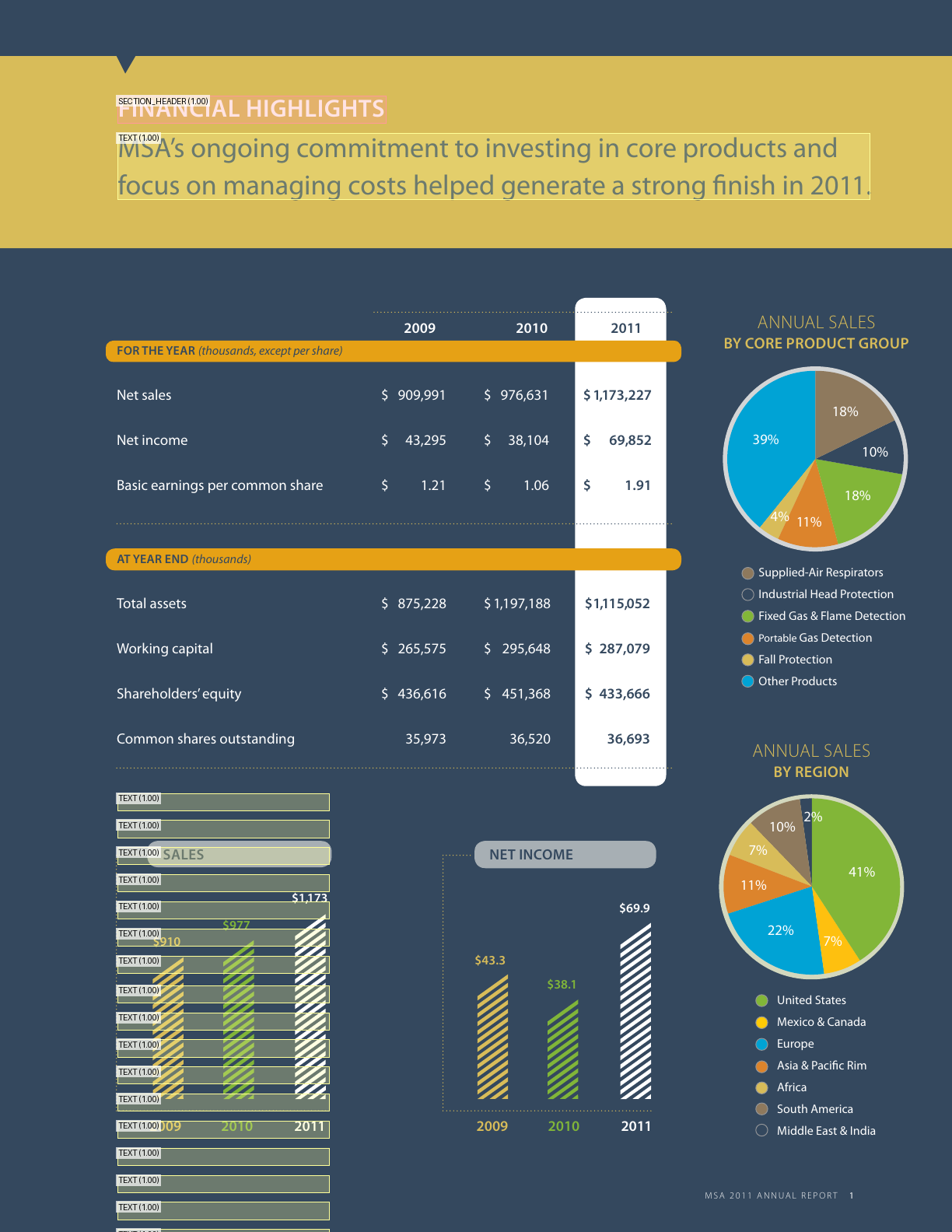} & 
    \includegraphics[width=0.31\columnwidth]{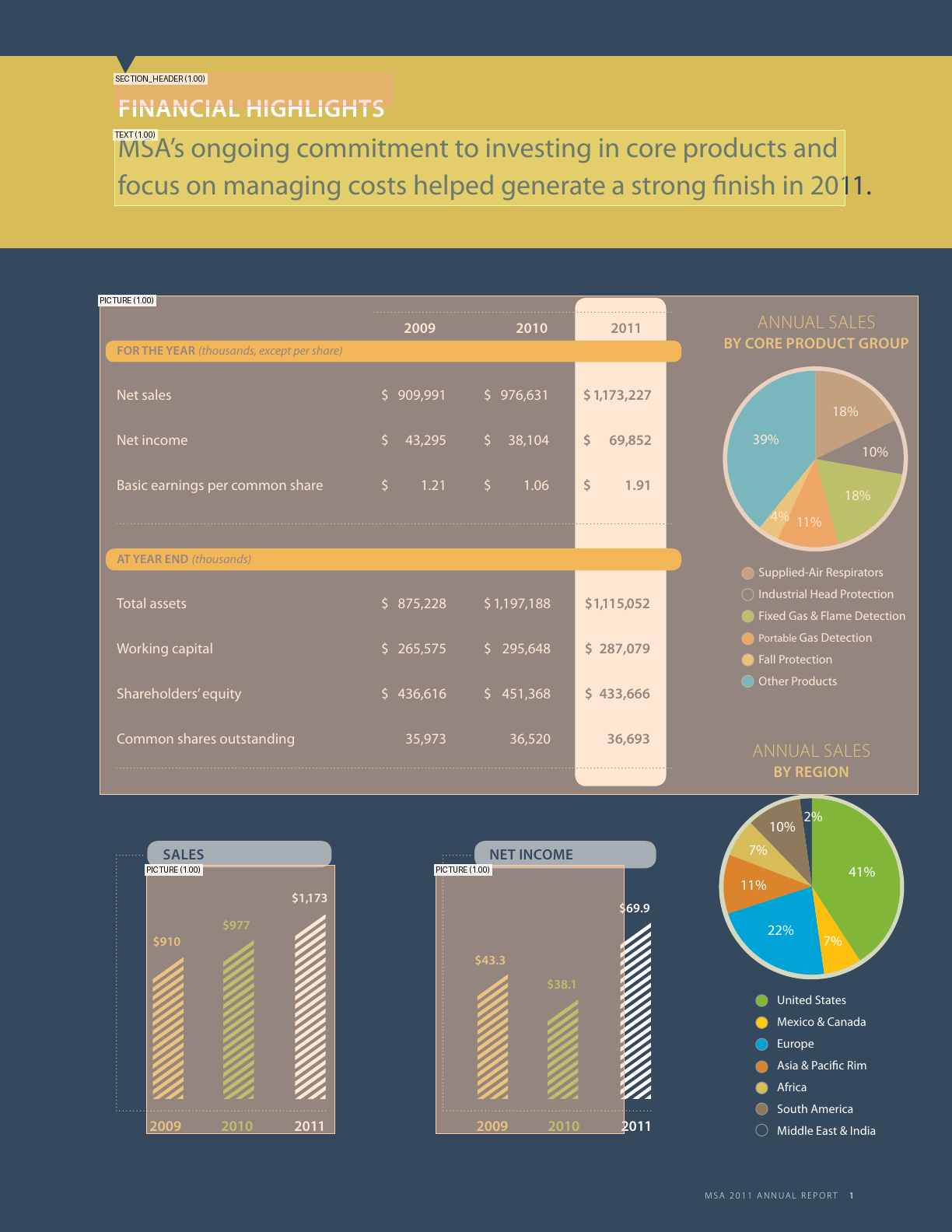} & 
    \rotatebox{90}{\tiny\ttfamily eae40317d83ba98d3e70427e6d7baff6} \\

\end{longtable}
    
\end{document}